%% file: main.tex
\documentclass{article}
\PassOptionsToPackage{round}{natbib}
\usepackage[final]{neurips_2024}

\usepackage[utf8]{inputenc} % allow utf-8 input
\usepackage[T1]{fontenc}    % use 8-bit T1 fonts
\usepackage{hyperref}       % hyperlinks
\usepackage{url}            % simple URL typesetting
\usepackage{booktabs}       % professional-quality tables
\usepackage{multirow,makecell}
\usepackage{graphicx}
\usepackage{caption}
\usepackage{subcaption}
\usepackage{algorithm}
\usepackage{algpseudocode}
\usepackage{mathtools}
\usepackage{amsmath}
\usepackage{amssymb}
\usepackage{amsthm}
\usepackage{amsfonts}
\usepackage{nicefrac}       % compact symbols for 1/2, etc.
\usepackage{microtype}      % microtypography
\usepackage{xcolor}         % colors
\usepackage[capitalise]{cleveref}

\hypersetup{colorlinks,linkcolor=red,citecolor=teal,urlcolor=blue}        % colors
\input{commands}
\newtheorem{theorem}{Theorem}
\newtheorem{assumption}{Assumption}
\newtheorem{lemma}{Lemma}
\newtheorem*{lemma*}{Lemma}
\crefname{assumption}{Assumption}{Assumptions}
\title{Improving robustness to corruptions with multiplicative weight perturbations}

\author{
  Trung Trinh$^1$ 
  \qquad
  Markus Heinonen$^1$
  \qquad
  Luigi Acerbi$^2$
  \qquad
  Samuel Kaski$^{1,3}$\\
  $^1$Department of Computer Science, Aalto University, Finland\\
  $^2$Department of Computer Science, University of Helsinki, Finland\\
  $^3$Department of Computer Science, University of Manchester, United Kingdom\\
  \texttt{\{trung.trinh, markus.o.heinonen, samuel.kaski\}@aalto.fi},\\ \texttt{luigi.acerbi@helsinki.fi}
}

\begin{document}

\maketitle

\begin{abstract}
   Deep neural networks (DNNs) excel on clean images but struggle with corrupted ones. Incorporating specific corruptions into the data augmentation pipeline can improve robustness to those corruptions but may harm performance on clean images and other types of distortion. In this paper, we introduce an alternative approach that improves the robustness of DNNs to a wide range of corruptions without compromising accuracy on clean images. We first demonstrate that input perturbations can be mimicked by multiplicative perturbations in the weight space. Leveraging this, we propose Data Augmentation via Multiplicative Perturbation (DAMP), a training method that optimizes DNNs under random multiplicative weight perturbations. We also examine the recently proposed Adaptive Sharpness-Aware Minimization (ASAM) and show that it optimizes DNNs under adversarial multiplicative weight perturbations. Experiments on image classification datasets (CIFAR-10/100, TinyImageNet and ImageNet) and neural network architectures (ResNet50, ViT-S/16, ViT-B/16) show that DAMP enhances model generalization performance in the presence of corruptions across different settings.
   Notably, DAMP is able to train a ViT-S/16 on ImageNet from scratch, reaching the top-1 error of $23.7\%$ which is comparable to ResNet50 without extensive data augmentations.\footnote{Our code is available at \url{https://github.com/trungtrinh44/DAMP}}
\end{abstract}

\section{Introduction}
Deep neural networks (DNNs) demonstrate impressive accuracy in computer vision tasks when evaluated on carefully curated and clean datasets. However, their performance significantly declines when test images are affected by natural distortions such as camera noise, changes in lighting and weather conditions, or image compression algorithms \citep{hendrycks2019benchmarking}. This drop in performance is problematic in production settings, where models inevitably encounter such perturbed inputs. Therefore, it is crucial to develop methods that produce reliable DNNs robust to common image corruptions, particularly for deployment in safety-critical systems \citep{amodei2016concrete}.

To enhance robustness against a specific corruption, one could simply include it in the data augmentation pipeline during training. However, this approach can diminish performance on clean images and reduce robustness to other types of corruptions \citep{geirhos2018generalisation}. More advanced data augmentation techniques \citep{cubuk2018autoaugment,hendrycks2019augmix,lopes2019improving} have been developed which effectively enhance corruption robustness without compromising accuracy on clean images. Nonetheless, a recent study by \citet{mintun2021on} has identified a new set of image corruptions to which models trained with these techniques remain vulnerable. Besides data augmentation, ensemble methods such as Deep ensembles and Bayesian neural networks have also been shown to improve generalization in the presence of corruptions \citep{lakshminarayanan2017simple,ovadia2019canyou,dusenberry20a,trinh22a}. However, the training and inference costs of these methods increase linearly with the number of ensemble members, rendering them less suitable for very large DNNs.

\begin{figure}[t]
   \centering
   \begin{subfigure}[b]{0.35\textwidth}
       \centering
       \includegraphics[width=\textwidth]{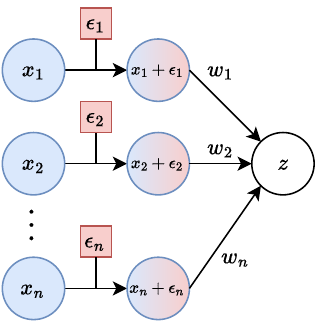}
       \caption{$z=\w^\top(\x+{\color{fig1red}\be})$}
       \label{fig:1a}
   \end{subfigure}
   \hfill
   \begin{subfigure}[b]{0.28\textwidth}
       \centering
       \includegraphics[width=\textwidth]{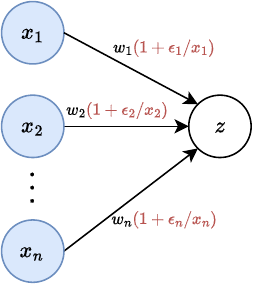}
       \caption{$z=(\w\circ{\color{fig1red} (1+\be/\x)})^\top\x$}
       \label{fig:1b}
   \end{subfigure}
   \hfill
   \begin{subfigure}[b]{0.28\textwidth}
       \centering
       \includegraphics[width=\textwidth]{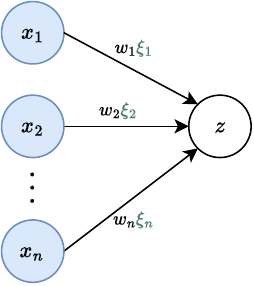}
       \caption{$z=(\w\circ{\color{fig1green}\bxi})^\top\x,\,\, {\color{fig1green}\bxi} \sim p({\color{fig1green}\bxi})$}
       \label{fig:1c}
   \end{subfigure}
      \caption{{\bf Depictions of a pre-activation neuron $z=\w^\top\x$ in the presence of (a) covariate shift ${\color{fig1red}\be}$, (b) a multiplicative weight perturbation (MWP) equivalent to ${\color{fig1red}\be}$, and (c) random MWPs ${\color{fig1green}\bxi}$.} $\circ$ denotes the Hadamard product. Figs. (a) and (b) show that for a covariate shift ${\color{fig1red}\be}$, one can always find an equivalent MWP. From this intuition, we propose to inject random MWPs ${\color{fig1green}\bxi}$ to the forward pass during training as shown in Fig. (c) to robustify a DNN to covariate shift.}
      \label{fig:1}
\end{figure}
\paragraph{Contributions} In this work, we show that simply perturbing weights with multiplicative random variables during training can significantly improve robustness to a wide range of corruptions. Our contributions are as follows:
\begin{itemize}
   \item We show in \cref{sec:damp} and \cref{fig:1} that the effects of input corruptions can be simulated during training via multiplicative weight perturbations.
   \item From this insight, we propose a new training algorithm called Data Augmentation via Multiplicative Perturbations (DAMP) which perturbs weights using multiplicative Gaussian random variables during training while having the same training cost as standard SGD.
   \item In \cref{sec:asam_mult}, we show a connection between adversarial multiplicative weight perturbations and Adaptive Sharpness-Aware Minimization (ASAM) \citep{kwon2021asam}.
   \item Through a rigorous empirical study in \cref{sec:evaluation}, we demonstrate that DAMP consistently improves generalization ability of DNNs under corruptions across different image classification datasets and model architectures.
   \item Notably, we demonstrate that DAMP can train a Vision Transformer (ViT) \citep{dosovitskiy2021an} from scratch on ImageNet, achieving similar accuracy to a ResNet50 \citep{he2016deep} in 200 epochs with only basic Inception-style preprocessing \citep{szegedy2016rethinking}. This is significant as ViT typically requires advanced training methods or sophisticated data augmentation to match ResNet50's performance when being trained on ImageNet from scratch \citep{chen2022when,beyer2022better}. We also show that DAMP can be combined with modern augmentation techniques such as MixUp \citep{zhang2018mixup} and RandAugment \citep{cubuk2020randaugment} to further improve robustness of neural networks.
\end{itemize}

\section{Data Augmentation via Multiplicative Perturbations}\label{sec:damp}
In this section, we demonstrate the equivalence between input corruptions and multiplicative weight perturbations (MWPs), as shown in \cref{fig:1}, motivating the use of MWPs for data augmentation.
\subsection{Problem setting}
Given a training data set $\S = \{(\x_k, y_k)\}_{k=1}^N \subseteq \X \times \Y$ drawn i.i.d.\ from the data distribution $\D$, we seek to learn a model that generalizes well on both clean and corrupted inputs.
We denote $\G$ as a set of functions whose each member $\g: \X \rightarrow \X$ represents an input corruption.
That is, for each $\x \in \X$, $\g(\x)$ is a corrupted version of $\x$.\footnote{For instance, if $\x$ is a clean image then $\g(\x)$ could be $\x$ corrupted by \emph{Gaussian noise}.}
We define $\g(\S) \defeq \{(\g(\x_k), y_k)\}_{k=1}^N$ as the training set corrupted by $\g$. 
We consider a DNN $\f: \X \rightarrow \Y$ parameterized by $\bo \in \cW$. Given a per-sample loss $\ell: \cW \times \X \times \Y \rightarrow \R_+$, the training loss is defined as the average loss over the samples $\L(\bo; \S) \defeq \frac{1}{N}\sum_{k=1}^N \ell(\bo,\x_k,y_k)$. 
Our goal is to find $\bo$ which minimizes:
\begin{equation}\label{eq:target_loss}
    \L(\bo; \G(\S)) \defeq \E_{\g \sim \G}[\L(\bo; \g(\S))]
\end{equation}
without knowing exactly the types of corruption contained in $\G$. This problem is crucial for the reliable deployment of DNNs, especially in safety-critical systems, since it is difficult to anticipate all potential types of corruption the model might encounter in production.
\subsection{Multiplicative weight perturbations simulate input corruptions}
\begin{figure}[t]
    \centering
    \includegraphics[width=.7\linewidth]{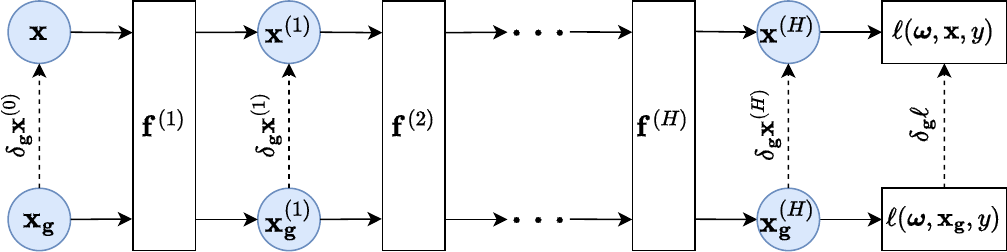}
    \caption{{\bf Depiction of how a corruption $\g$ affects the output of a DNN.} Here $\x_\g=\g(\x)$. The corruption $\g$ creates a shift $\bd_\g\x^{(0)}=\x_\g-\x$ in the input $\x$, which propagates into shifts $\bd_\g\x^{(h)}$ in the output of each layer. This will eventually cause a shift in the loss $\bd_\g \ell$. This figure explains why the model performance tends to degrade under corruption.}
    \label{fig:fig2}
\end{figure}
To address the problem above, we make two key assumptions about the corruptions in $\G$:
\begin{assumption}[Bounded corruption]\label{assumption:bounded_g}
    For each corruption function $\g: \X \rightarrow \X$ in $\G$, there exists a constant $M > 0$ such that $\|\g(\x)-\x\|_2 \leq M$ for all $\x \in \X$.
\end{assumption}
\begin{assumption}[Transferable robustness]\label{assumption:transferable}
    A model's robustness to corruptions in $\G$ can be indirectly enhanced by improving its resilience to a more easily simulated set of input perturbations.
\end{assumption}
\cref{assumption:bounded_g} implies that the corrupted versions of an input $\x$ must be constrained within a bounded neighborhood of $\x$ in the input space. \cref{assumption:transferable} is corroborated by \citet{rusak2020simple}, who demonstrated that distorting training images with Gaussian noise improves a DNN's performance against various types of corruption. We further validate this observation for corruptions beyond Gaussian noise in \cref{sec:damp_vs_corr}. However, \cref{sec:damp_vs_corr} also reveals that using corruptions as data augmentation degrades model performance on clean images. Consequently, we need to identify a method that efficiently simulates diverse input corruptions during training, thereby robustifying a DNN against a wide range of corruptions without compromising its performance on clean inputs.

One such method involves injecting random multiplicative weight perturbations (MWPs) into the forward pass of DNNs during training. The intuition behind this approach is illustrated in \cref{fig:1}. Essentially, for a pre-activated neuron $z=\w^\top \x$ in a DNN, given a corruption causing a covariate shift $\be$ in the input $\x$, \cref{fig:1a,fig:1b} show that one can always find an equivalent MWP $\bxi(\be,\x)$:
\begin{equation}
    z = \w^\top(\x + \be) = (\w \circ \bxi(\be,\x))^\top \x, \quad \bxi(\be,\x)=1+\be/\x
\end{equation}
where $\circ$ denotes the Hadamard product. This observation suggests that input corruptions can be simulated during training by injecting random MWPs into the forward pass, as depicted in \cref{fig:1c}, resulting in a model more robust to corruption. We thus move the problem of simulating corruptions from the input space to the weight space.

Here we provide theoretical arguments supporting the usage of MWPs to robustify DNNs. To this end, we study how corruption affects training loss.
We consider a feedforward neural network $\f(\x; \bo)$ of depth $H$ parameterized by $\bo = \{\W^{(h)}\}_{h=1}^H \in \cW$, which we define recursively as follows:
\begin{align}
    \f^{(0)}(\x)\defeq\x, \quad \z^{(h)}(\x) \defeq \W^{(h)} \f^{(h-1)}(\x), \quad \f^{(h)}(\x) \defeq \bs^{(h)}(\z^{(h)}(\x)), \quad \forall h=1,\dots,H
\end{align}
where $\f(\x;\bo)\defeq\f^{(H)}(\x)$ and $\bs^{(h)}$ is the non-linear activation of layer $h$. 
For brevity, we use $\x^{(h)}$ and $\x^{(h)}_\g$ as shorthand notations for $\f^{(h)}(\x)$ and $\f^{(h)}(\g(\x))$ respectively.
Given a corruption function $\g$, \cref{fig:fig2} shows that $\g$ creates a covariate shift $\bd_\g\x^{(0)} \defeq \x_\g^{(0)}-\x^{(0)}$ in the input $\x$ leading to shifts $\bd_\g\x^{(h)} \defeq \x^{(h)}_\g-\x^{(h)}$ in the output of each layer.
This will eventually cause a shift in the per-sample loss $\bd_\g \ell(\bo, \x, y) \defeq \ell(\bo, \x_\g, y) - \ell(\bo, \x, y)$.
The following lemma characterizes the connection between $\bd_\g \ell(\bo, \x, y)$ and $\bd_\g\x^{(h)}$:
\begin{lemma}\label{lemma:per_sample_loss_bound}
    For all $h=1,\dots,H$ and for all $\x \in \X$, there exists a scalar $C_\g^{(h)}(\x) > 0$ such that:
    \begin{equation}\label{eq:lemma_1}
        {\color{eqpurple}\bd_\g \ell(\bo, \x, y)} \leq {\color{eqred}{\left\langle \nabla_{\z^{(h+1)}} \ell(\bo, \x, y) \otimes \bd_\g\x^{(h)}, \W^{(h+1)} \right\rangle}_F} + {\color{eqblue}\frac{C_\g^{(h)}(\x)}{2}\|\W^{(h)}\|_F^2}
    \end{equation}
\end{lemma}
Here $\otimes$ denotes the outer product of two vectors, $\langle \cdot, \cdot \rangle_F$ denotes the Frobenius inner product of two matrices of the same dimension, $\|\cdot\|_F$ is the Frobenius norm, and $\nabla_{\z^{(h)}}\ell(\bo,\x,y)$ is the Jacobian of the per-sample loss with respect to the pre-activation output $\z^{(h)}(\x)$ at layer $h$. To prove \cref{lemma:per_sample_loss_bound}, we use \cref{assumption:bounded_g} and the following assumption about the loss function:
\begin{assumption}[Lipschitz-continuous objective input gradients]\label{assumption:lipschitz_grad}
    The input gradient of the per-sample loss $\nabla_{\x} \ell(\bo,\x,y)$ is Lipschitz continuous.
\end{assumption}
\cref{assumption:lipschitz_grad} allows us to define a quadratic bound of the loss function using a second-order Taylor expansion.
The proof of \cref{lemma:per_sample_loss_bound} is provided in \cref{sec:lemma_1_proof}. Using \cref{lemma:per_sample_loss_bound}, we prove \cref{theorem:bounded_loss}, which bounds the training loss in the presence of corruptions using the training loss under multiplicative perturbations in the weight space:
\begin{theorem}\label{theorem:bounded_loss}
    For a function $\g: \X \rightarrow \X$ satisfying \cref{assumption:bounded_g} and a loss function $\L$ satisfying \cref{assumption:lipschitz_grad}, there exists $\bxi_\g \in \cW$ and $C_\g > 0$ such that:
    \begin{equation}
        {\color{eqpurple} \L(\bo; \g(\S))} \leq {\color{eqred}\L(\bo \circ \bxi_\g; \S)} + {\color{eqblue} \frac{C_\g}{2}\|\bo\|_F^2} \label{eq:bounded_loss}
    \end{equation}
\end{theorem}
We provide the proof of \cref{theorem:bounded_loss} in \cref{sec:theorem_1_proof}. This theorem establishes an upper bound for the target loss in \cref{eq:target_loss}:
\begin{equation}
    {\color{eqpurple}\L(\bo;\G(\S))} \leq \E_{\g \sim \G}\left[ {\color{eqred}\L(\bo \circ \bxi_\g; \S)} + {\color{eqblue}\frac{C_\g}{2} \|\bo\|_F^2} \right] 
\end{equation}
This bound implies that training a DNN using the following loss function:
\begin{equation}\label{eq:mult_aux_loss}
    {\color{eqpurple}\L_{\bXi}(\bo; \S)} \defeq \E_{\bxi \sim \bXi}\left[{\color{eqred}\L(\bo \circ \bxi; \S)}\right] + {\color{eqblue}\frac{\lambda}{2}\|\bo\|_F^2}
\end{equation}
where the expected loss is taken with respect to a distribution $\bXi$ of random MWPs $\bxi$, will minimize the upper bound of the loss $\color{eqpurple}\L(\bo;\hat{\G}(\S))$ of a hypothetical set of corruptions $\hat{\G}$ simulated by $\bxi \sim \bXi$. This approach results in a model robust to these simulated corruptions, which, according to \cref{assumption:transferable}, could indirectly improve robustness to corruptions in $\G$.
\input{damp}
We note that the second term in \cref{eq:mult_aux_loss} is the $L_2$-regularization commonly used in optimizing DNNs. Based on this proxy loss, we propose \cref{alg:damp} which minimizes the objective function in \cref{eq:mult_aux_loss} when $\bXi$ is an isotropic Gaussian distribution $\N(\boldsymbol{1}, \sigma^2\mathbf{I})$. We call this algorithm Data Augmentation via Multiplicative Perturbations (DAMP), as it uses random MWPs during training to simulate input corruptions, which can be viewed as data augmentations.
\paragraph{Remark} The standard method to calculate the expected loss in \cref{eq:mult_aux_loss}, which lacks a closed-form solution, is the Monte Carlo (MC) approximation. However, the training cost of this approach scales linearly with the number of MC samples. To match the training cost of standard SGD, \cref{alg:damp} divides each data batch into $M$ equal-sized sub-batches (\cref{alg:divide_step}) and calculates the loss on each sub-batch with different multiplicative noises from the noise distribution $\bXi$ (\crefrange{alg:separated_grad_start}{alg:separate_grad_end}). The final gradient is obtained by averaging the sub-batch gradients (\cref{alg:avg_grad}). \cref{alg:damp} is thus suitable for data parallelism in multi-GPU training, where the data batch is evenly distributed across $M > 1$ GPUs. Compared to SGD, \cref{alg:damp} requires only two additional operations: generating Gaussian samples and point-wise multiplication, both of which have negligible computational costs. In our experiments, we found that both SGD and DAMP had similar training times.

\section{Adaptive Sharpness-Aware Minimization optimizes DNNs under adversarial multiplicative weight perturbations}\label{sec:asam_mult}
In this section, we demonstrate that optimizing DNNs with adversarial MWPs follows a similar update rule to Adaptive Sharpness-Aware Minimization (ASAM) \citep{kwon2021asam}. We first provide a brief description of ASAM and its predecessor Sharpness-Aware Minimization (SAM) \citep{foret2021sharpnessaware}:
\paragraph{SAM} Motivated by previous findings that wide optima tend to generalize better than sharp ones \citep{keskar2017on,Jiang2020Fantastic}, SAM regularizes the sharpness of an optimum by solving the following minimax optimization:
\begin{equation}
    \min_{\bo} \max_{\|\bxi\|_2 \leq \rho} {\color{eqred}\L(\bo + \bxi; \S)} + {\color{eqblue}\frac{\lambda}{2}\|\bo\|_F^2}
\end{equation}
which can be interpreted as optimizing DNNs under adversarial additive weight perturbations. To efficiently solve this problem, \citet{foret2021sharpnessaware} devise a two-step procedure for each iteration $t$:
\begin{align}
    \bxi^{(t)} = \rho \frac{\nabla_{\bo} \L(\bo^{(t)}; \S)}{\left|\left|\nabla_{\bo} \L(\bo^{(t)}; \S)\right|\right|_2},
    \qquad \bo^{(t+1)} = \bo^{(t)} - \eta_t \left({\color{eqred}\nabla_{\bo}\L(\bo^{(t)}+\bxi^{(t)}; \S)} + {\color{eqblue}\lambda \bo^{(t)}}\right)
\end{align}
where $\eta_t$ is the learning rate. Each iteration of SAM thus takes twice as long to run than SGD.
\paragraph{ASAM} \citet{kwon2021asam} note that SAM attempts to minimize the maximum loss over a rigid sphere of radius $\rho$ around an optimum, which is not suitable for ReLU networks since their parameters can be freely re-scaled without affecting the outputs. The authors thus propose ASAM as an alternative optimization problem to SAM which regularizes the \emph{adaptive sharpness} of an optimum:
\begin{equation}\label{eq:asam_def}
    \min_{\bo} \max_{\|T_{\bo}^{-1}\bxi\|_2 \leq \rho} {\color{eqred}\L(\bo + \bxi; \S)} + {\color{eqblue}\frac{\lambda}{2}\|\bo\|_F^2}
\end{equation}
where $T_{\bo}$ is an invertible linear operator used to reshape the perturbation region (so that it is not necessarily a sphere as in SAM). \citet{kwon2021asam} found that $T_{\bo} = |\bo|$ produced the best results. Solving \cref{eq:asam_def} in this case leads to the following two-step procedure for each iteration $t$:
\begin{align}\label{eq:asam_update}
    \widehat{\bxi}^{(t)} = \rho \frac{\left(\bo^{(t)}\right)^2 \circ \nabla_{\bo} \L(\bo^{(t)}; \S)}{\left|\left| \bo^{(t)} \circ \nabla_{\bo} \L(\bo^{(t)}; \S)\right|\right|_2}, 
    \quad \bo^{(t+1)} = \bo^{(t)} - \eta_t \left({\color{eqred}\nabla_{\bo}\L(\bo^{(t)}+\widehat{\bxi}^{(t)}; \S)} + {\color{eqblue}\lambda \bo^{(t)}}\right)
\end{align}
Similar to SAM, each iteration of ASAM also takes twice as long to run than SGD.
\paragraph{ASAM and adversarial multiplicative perturbations} \cref{alg:damp} minimizes the expected loss in \cref{eq:mult_aux_loss}.
Instead, we could minimize the loss under the adversarial MWP:
\begin{equation}\label{eq:l_max}
    {\color{eqpurple}\Lmax(\bo; \S)} \defeq \max_{\|\bxi\|_2 \leq \rho} {\color{eqred}\L(\bo + \bo \circ \bxi; \S)} + {\color{eqblue}\frac{\lambda}{2}\|\bo\|_F^2}
\end{equation}
Following \citet{foret2021sharpnessaware}, we solve this optimization problem by using a first-order Taylor expansion of $\color{eqred}\L(\bo + \bo \circ \bxi; \S)$ to find an approximate solution of the inner maximization:
\begin{align}
    \argmax_{\|\bxi\|_2 \leq \rho} {\color{eqred}\L(\bo+\bo\circ\bxi;\S)} \approx \argmax_{\|\bxi\|_2 \leq \rho} \L(\bo; \S) + \left\langle \bo\circ\bxi, \nabla_{\bo} \L(\bo;\S)  \right\rangle
\end{align}
The maximizer of the Taylor expansion is:
\begin{equation}\label{eq:bxi_argmax}
    \widehat{\bxi}(\bo)  = \rho \frac{\bo\circ\nabla_{\bo} \L(\bo;\S)}{\|\bo\circ\nabla_{\bo} \L(\bo;\S)\|_2}
\end{equation}
Subtituting back into \cref{eq:l_max} and differentiating, we get:
\begin{align}
    {\color{eqpurple}\nabla_{\bo}\Lmax(\bo; \S)} &\approx {\color{eqred}\nabla_{\bo} \L(\widehat{\bo}; \S)} + {\color{eqblue}\lambda\bo} = {\color{eqred}\nabla_{\bo}\widehat{\bo} \cdot \nabla_{\widehat{\bo}}\L(\widehat{\bo};\S)} + {\color{eqblue}\lambda\bo} \\
    &= {\color{eqred}\nabla_{\widehat{\bo}}\L(\widehat{\bo};\S)} + {\color[HTML]{FF898C}\nabla_{\bo}\left(\bo\circ\widehat{\bxi}(\bo)\right)\cdot\nabla_{\widehat{\bo}}\L(\widehat{\bo};\S)}  + {\color{eqblue}\lambda\bo} \label{eq:l_max_full_grad}
\end{align}
where $\widehat{\bo}$ is the perturbed weight:
\begin{equation}
    \widehat{\bo} = \bo + \bo \circ \widehat{\bxi}(\bo) =  \bo + \rho\frac{\bo^2 \circ \nabla_{\bo} \L(\bo;\S)}{\|\bo \circ \nabla_{\bo} \L(\bo;\S)\|_2}
\end{equation}
Similar to \cite{foret2021sharpnessaware}, we omit the second summand in \cref{eq:l_max_full_grad} for efficiency, as it requires calculating the Hessian of the loss. We then arrive at the gradient formula in the update rule of ASAM in \cref{eq:asam_update}. We have thus established a connection between ASAM and adversarial MWPs.

\section{Empirical evaluation} \label{sec:evaluation}
In this section, we assess the corruption robustness of DAMP and ASAM in image classification tasks. We conduct experiments using the CIFAR-10/100 \citep{krizhevsky2009cifar}, TinyImageNet \citep{Le2015TinyIV}, and ImageNet \citep{deng2009} datasets. For evaluation on corrupted images, we utilize the CIFAR-10/100-C, TinyImageNet-C, and ImageNet-C datasets provided by \citet{hendrycks2019benchmarking}, as well as ImageNet-$\Cbar$ \citep{mintun2021on}, ImageNet-D \citep{imagenet-d}, ImageNet-A \citep{imagenet-a}, ImageNet-Sketch \citep{imagenet-sketch}, ImageNet-\{Drawing, Cartoon\} \citep{imagenet-cartoon-drawing}, and ImageNet-Hard \citep{taesiri2023zoom} datasets, which encapsulate a wide range of corruptions. Detail descriptions of these datasets are provided in \cref{sec:corruption_datasets}. We further evaluate the models on adversarial examples generated by the Fast Gradient Sign Method (FGSM) \citep{fgsm}. In terms of architectures, we use ResNet18 \citep{he2016deep} for CIFAR-10/100, PreActResNet18 \citep{he2016identity} for TinyImageNet, ResNet50 \citep{he2016deep}, ViT-S/16, and ViT-B/16 \citep{dosovitskiy2021an} for ImageNet. We ran all experiments on a single machine with 8 Nvidia V100 GPUs. \cref{sec:training_detail} includes detailed information for each experiment.

\subsection{Comparing DAMP to directly using corruptions as augmentations}\label{sec:damp_vs_corr}
\begin{figure}[t]
    \centering
    \includegraphics[width=.8\linewidth]{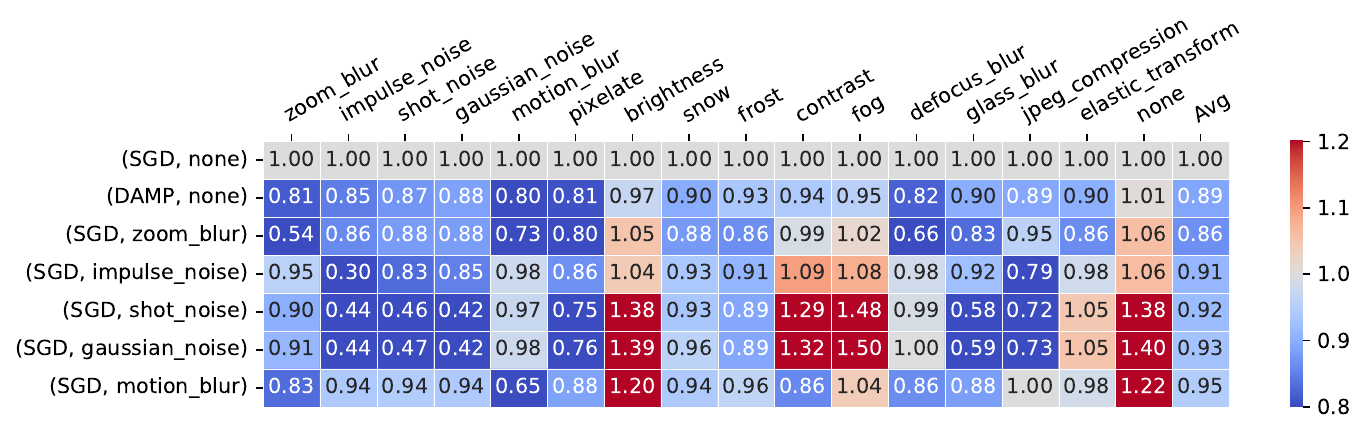}
    \caption{{\bf DAMP improves robustness to all corruptions while preserving accuracy on clean images.} Results of ResNet18/CIFAR-100 experiments averaged over 5 seeds. The heatmap shows $\CE^{f}_c$ described in \cref{eq:ce_c_f} (lower is better), where each row corresponds to a tuple of training (\texttt{method}, \texttt{corruption}), while each column corresponds to the test corruption. The \texttt{Avg} column shows the average of the results of the previous columns. \texttt{none} indicates no corruption. We use the models trained under the \texttt{SGD/none} setting (first row) as baselines to calculate the $\CE^{f}_c$. The last five rows are the 5 best training corruptions ranked by the results in the \texttt{Avg} column.}
    \label{fig:cifar100_corr_experiment_results}
\end{figure}
In this section, we compare the corruption robustness of DNNs trained using DAMP with those trained directly on corrupted images. To train models on corrupted images, we utilize \cref{alg:corr_training} described in the Appendix. For a given target corruption $\g$, \cref{alg:corr_training} randomly selects half the images in each training batch and applies $\g$ to them. This random selection process enhances the final model's robustness to the target corruption while maintaining its accuracy on clean images. We use the \texttt{imagecorruptions} library \citep{michaelis2019dragon} to apply the corruptions during training.
\paragraph{Evaluation metric} We use the corruption error $\CE_c^f$ \citep{hendrycks2019benchmarking} which measures the predictive error of classifier $f$ in the presence of corruption $c$. Denote $E^f_{s,c}$ as the error of classifier $f$ under corruption $c$ with corruption severity $s$, the corruption error $\CE_c^f$ is defined as:
\begin{equation}\label{eq:ce_c_f}
    \CE_c^f = \biggl(\sum_{s=1}^5 E^f_{s,c}\biggr)\bigg/\biggl(\sum_{s=1}^5 E^{f_{\text{baseline}}}_{s,c}\biggr)
\end{equation}
For this metric, lower is better. Here $f_{\text{baseline}}$ is a baseline classifier whose usage is to make the error more comparable between corruptions as some corruptions can be more challenging than others \citep{hendrycks2019benchmarking}. For each experiment setting, we use the model trained by SGD without corruptions as $f_{\text{baseline}}$.
\paragraph{Results} We visualize the results for the ResNet18/CIFAR-100 setting in \cref{fig:cifar100_corr_experiment_results}. The results for the ResNet18/CIFAR-10 and PreActResNet18/TinyImageNet settings are presented in \cref{fig:cifar10_corr_experiment_results,fig:tinyimagenet_corr_experiment_results} in the Appendix. \cref{fig:cifar100_corr_experiment_results,fig:cifar10_corr_experiment_results,fig:tinyimagenet_corr_experiment_results} demonstrate that DAMP improves predictive accuracy over plain SGD across all corruptions without compromising accuracy on clean images. Although \cref{fig:cifar100_corr_experiment_results} indicates that including \texttt{zoom\_blur} as an augmentation when training ResNet18 on CIFAR-100 yields better results than DAMP on average, it also reduces accuracy on clean images and the \texttt{brightness} corruption. Overall, these figures show that incorporating a specific corruption as data augmentation during training enhances robustness to that particular corruption but may reduce performance on clean images and other corruptions. In contrast, DAMP consistently improves robustness across all corruptions. Notably, DAMP even enhances accuracy on clean images in the PreActResNet18/TinyImageNet setting, as shown in \cref{fig:tinyimagenet_corr_experiment_results}.

\subsection{Comparing DAMP to random additive perturbations}\label{sec:damp_vs_daap}
In this section, we investigate whether additive weight perturbations can also enhance corruption robustness.
To this end, we compare DAMP with its variant, Data Augmentation via Additive Perturbations (DAAP). Unlike DAMP, DAAP perturbs weights during training with random additive Gaussian noises centered at $0$, as detailed in \cref{alg:daap} in the Appendix. \cref{fig:daap_vs_damp} in the Appendix presents the results of DAMP and DAAP under different noise standard deviations, alongside standard SGD. Overall, \cref{fig:daap_vs_damp} shows that across different experimental settings, the corruption robustness of DAAP is only slightly better than SGD and is worse than DAMP. Therefore, we conclude that MWPs are better than their additive counterparts at improving robustness to corruptions.

\subsection{Benchmark results}
\begin{figure}[t]
    \centering
    \includegraphics[width=.8\linewidth]{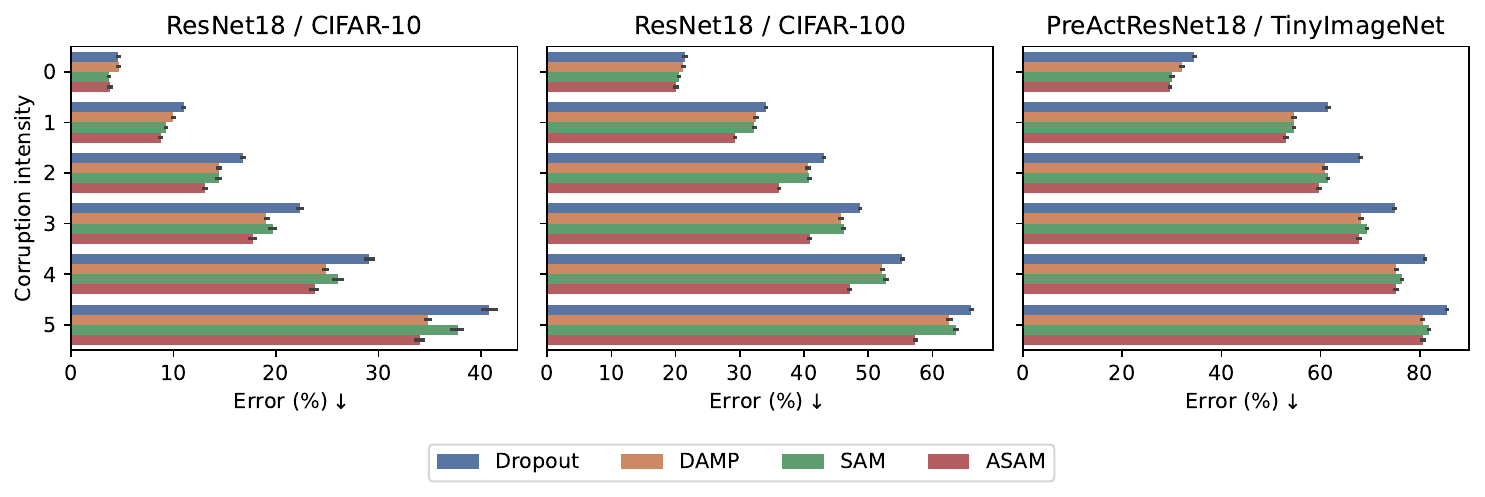}
    \caption{{\bf DAMP surpasses SAM on corrupted images in most cases, despite requiring only half the training cost.} We report the predictive errors (lower is better) averaged over 5 seeds. A severity level of $0$ indicates no corruption. We use the same number of epochs for all methods.}
    \label{fig:small_benchmark}
\end{figure}
In this section, we compare DAMP with Dropout \citep{srivastava14a}, SAM \citep{foret2021sharpnessaware}, and ASAM \citep{kwon2021asam}. For SAM and ASAM, we optimize the neighborhood size $\rho$ by using $10\%$ of the training set as a validation set. Similarly, we adjust the noise standard deviation $\sigma$ for DAMP and the drop rate $p$ for Dropout following the same procedure. For hyperparameters and additional training details, please refer to \cref{sec:training_detail}.
\paragraph{CIFAR-10/100 and TinyImageNet.} \cref{fig:small_benchmark} visualizes the predictive errors of DAMP and the baseline methods on CIFAR-10/100 and TinyImageNet, with all methods trained for the same number of epochs. It demonstrates that DAMP consistently outperforms Dropout across various datasets and corruption severities, despite having the same training cost. Notably, DAMP outperforms SAM under most corruption scenarios, even though SAM takes twice as long to train and has higher accuracy on clean images. Additionally, DAMP improves accuracy on clean images over Dropout on CIFAR-100 and TinyImageNet. Finally, ASAM consistently surpasses other methods on both clean and corrupted images, as it employs adversarial MWPs (\cref{sec:asam_mult}). However, like SAM, each ASAM experiment takes twice as long as DAMP given the same epoch counts.
\input{resnet50_imagenet_table}
\paragraph{ResNet50 / ImageNet} \cref{tab:resnet50_imagenet} presents the predictive errors for the ResNet50 / ImageNet setting on a variety of corruption test sets. It shows that DAMP consistently outperforms the baselines in most corruption scenarios and on average, despite having half the training cost of SAM and ASAM.
\input{vits16_imagenet_table}
\paragraph{ViT-S16 / ImageNet / Basic augmentations}\cref{tab:vits16_imagenet} presents the predictive errors for the ViT-S16 / ImageNet setting, using the training setup from \citet{beyer2022better} but with only basic Inception-style preprocessing \citep{szegedy2016rethinking}. Remarkably, DAMP can train ViT-S16 from scratch in 200 epochs to match ResNet50's accuracy without advanced data augmentation. This is significant as ViT typically requires either extensive pretraining \citep{dosovitskiy2021an}, comprehensive data augmentation \citep{beyer2022better}, sophisticated training techniques \citep{chen2022when}, or modifications to the original architecture \citep{yuan2021tokens} to perform well on ImageNet. Additionally, DAMP consistently ranks in the top 2 for corruption robustness across various test settings and has the best corruption robustness on average (last column). Comparing \cref{tab:resnet50_imagenet,tab:vits16_imagenet} reveals that ViT-S16 is more robust to corruptions than ResNet50 when both have similar performance on clean images.
\input{vit_imagenet_table_advanceaug}
\paragraph{ViT / ImageNet / Advanced augmentations}\cref{tab:vit_aug} presents the predictive errors of ViT-S16 and ViT-B16 on ImageNet with MixUp \citep{zhang2018mixup} and RandAugment \citep{cubuk2020randaugment}. These results indicate that DAMP can be combined with modern augmentation techniques to further improve robustness. Furthermore, using DAMP to train a larger model (ViT-B16) yields better results than using SAM/ASAM to train a smaller model (ViT-S16), given the same amount of training time.

\section{Related works}
\paragraph{Dropout} Perhaps most relevant to our method is Dropout \citep{srivastava14a} and its many variants, such as DropConnect \citep{pmlr-v28-wan13} and Variational Dropout \citep{kingma2015variational}. These methods can be viewed as DAMP where the noise distribution $\bXi$ is a structured multivariate Bernoulli distribution. For instance, Dropout multiplies all the weights connecting to a node with a binary random variable $p \sim \text{Bernoulli}(\rho)$. While the main motivation of these Dropout methods is to prevent co-adaptations of neurons to improve generalization on clean data, the motivation of DAMP is to improve robustness to input corruptions without harming accuracy on clean data. Nonetheless, our experiments show that DAMP can improve generalization on clean data in certain scenarios, such as PreActResNet18/TinyImageNet and ViT-S16/ImageNet.
\paragraph{Ensemble methods} Ensemble methods, such as Deep ensembles \citep{lakshminarayanan2017simple} and Bayesian neural networks (BNNs) \citep{graves2011,blundell2015weight,pmlr-v48-gal16,louizos2017multiplicative,izmailov2021dangers,trinh22a}, have been explored as effective defenses against corruptions. \citet{ovadia2019canyou} benchmarked some of these methods, demonstrating that they are more robust to corruptions compared to a single model. However, the training and inference costs of these methods increase linearly with the number of ensemble members, making them inefficient for use with very large DNNs.
\paragraph{Data augmentation} Data augmentations aim at enhancing robustness include AugMix \citep{hendrycks2019augmix}, which combines common image transformations; Patch Gaussian \citep{lopes2019improving}, which applies Gaussian noise to square patches; ANT \citep{rusak2020simple}, which uses adversarially learned noise distributions for augmentation; and AutoAugment \citep{cubuk2018autoaugment}, which learns augmentation policies directly from the training data. These methods have been demonstrated to improve robustness to the corruptions in ImageNet-C \citep{hendrycks2019benchmarking}. \citet{mintun2021on} attribute the success of these methods to the fact that they generate augmented images perceptually similar to the corruptions in ImageNet-C and propose ImageNet-$\Cbar$, a test set of 10 new corruptions that are challenging to models trained by these augmentation methods.
\paragraph{Test-time adaptations via BatchNorm} One effective approach to using unlabelled data to improve corruption robustness is to keep BatchNorm \citep{batchnorm2015} on at test-time to adapt the batch statistics to the corrupted test data \citep{li2016revisiting,nado2020evaluating,schneider2020improving,benz2021revisiting}. A major drawback is that this approach cannot be used with BatchNorm-free architectures, such as Vision Transformer \citep{dosovitskiy2021an}.

\section{Conclusion}\label{sec:conclusion}
In this work, we demonstrate that MWPs improve robustness of DNNs to a wide range of input corruptions. We introduce DAMP, a simple training algorithm that perturbs weights during training with random multiplicative noise while maintaining the same training cost as standard SGD. We further show that ASAM \citep{kwon2021asam} can be viewed as optimizing DNNs under adversarial MWPs. Our experiments show that both DAMP and ASAM indeed produce models that are robust to corruptions. DAMP is also shown to improve sample efficiency of Vision Transformer, allowing it to achieve comparable performance to ResNet50 on medium size datasets such as ImageNet without extensive data augmentations. Additionally, DAMP can be used in conjunction with modern augmentation techniques such as MixUp and RandAugment to further boost robustness. As DAMP is domain-agnostic, one future direction is to explore its effectiveness in domains other than computer vision, such as natural language processing and reinforcement learning. Another direction is to explore alternative noise distributions to the Gaussian distribution used in our work.
\paragraph{Limitations} Here we outline some limitations of this work. First, the proof of \cref{theorem:bounded_loss} assumes a simple feedforward neural network, thus it does not take into accounts modern DNN's components such as normalization layers and attentions. Second, we only explored random Gaussian multiplicative perturbations, and there are likely more sophisticated noise distributions that could further boost corruption robustness.

\section*{Broader Impacts}
Our paper introduces a new training method for neural networks that improves their robustness to input corruptions.
Therefore, we believe that our work contributes towards making deep leading models safer and more reliable to use in real-world applications, especially those that are safety-critical.
However, as with other methods that improve robustness, our method could also be improperly used in applications that negatively impact society, such as making mass surveillance systems more accurate and harder to fool.
To this end, we hope that practitioners carefully consider issues regarding fairness, bias and other potentially harmful societal impacts when designing deep learning applications.

\section*{Acknowledgments}
This work was supported by the Research Council of Finland (Flagship programme: Finnish Center for Artificial Intelligence FCAI and decision no. 359567, 345604 and 341763), ELISE Networks of Excellence Centres (EU Horizon: 2020 grant agreement 951847) and UKRI Turing AI World-Leading Researcher Fellowship (EP/W002973/1). We acknowledge the computational resources provided by Aalto Science-IT project and CSC-IT Center for Science, Finland.

\bibliographystyle{unsrtnat}
\bibliography{ref}
\newpage

\appendix
\input{appendix}

\newpage
\input{checklist}

\end{document}

%% file: commands.tex
\def\x{\mathbf{x}}
\def\S{\mathcal{S}}
\def\D{\mathcal{D}}

\def\bo{\boldsymbol{\omega}}

\def\X{\mathcal{X}}
\def\Y{\mathcal{Y}}
\def\R{\mathbb{R}}
\def\y{\mathbf{y}}
\def\L{\mathcal{L}}
\def\N{\mathcal{N}}
\def\E{\mathbb{E}}
\def\g{\mathbf{g}}
\def\G{\mathcal{G}}
\def\w{\mathbf{w}}
\def\be{\boldsymbol{\epsilon}}
\def\defeq{\vcentcolon=}
\def\f{\mathbf{f}}

\def\bs{\boldsymbol{\sigma}}
\def\W{\mathbf{W}}
\def\cW{\mathcal{W}}
\def\z{\mathbf{z}}
\def\bd{\boldsymbol{\delta}}

\def\bxi{\boldsymbol{\xi}}
\def\bXi{\boldsymbol{\Xi}}

\def\B{\mathcal{B}}
\def\Lmax{\L_{\mathrm{max}}}
\DeclareMathOperator*{\argmax}{arg\,max}

\def\CE{\mathrm{CE}}
\def\Cbar{\overline{\text{C}}}
\definecolor{fig1red}{HTML}{B85450}
\definecolor{fig1green}{HTML}{4A7A72}
\definecolor{mydarkblue}{rgb}{0,0.08,0.45}
\definecolor{eqred}{HTML}{FE433C}
\definecolor{eqblue}{HTML}{0995EF}
\definecolor{eqpurple}{HTML}{A224AD}

%% file: damp.tex
\begin{algorithm}[t]
    \caption{DAMP: Data Augmentation via Multiplicative Perturbations}\label{alg:damp}
 \begin{algorithmic}[1]
    \State {\bfseries Input:} training data $\S = \{(\x_k, y_k)\}_{k=1}^N$, a neural network $\f(\cdot; \bo)$ parameterized by $\bo \in \R^P$, number of iterations $T$, step sizes $\{\eta_t\}_{t=1}^T$, number of sub-batch $M$, batch size $B$ divisible by $M$, a noise distribution $\bXi = \N(\mathbf{1}, \sigma^2\mathbf{I}_P)$, weight decay coefficient $\lambda$, a loss function $\L: \R^P \rightarrow \R_+$.
    \State {\bfseries Output:} Optimized parameter $\bo^{(T)}$.
    \State Initialize parameter $\bo^{(0)}$.
    \For{$t=1$ {\bfseries to} $T$}
    \State Draw a mini-batch $\B=\{(\x_b, y_b)\}_{b=1}^B \sim \S$.
    \State Divide the mini-batch into $M$ \emph{disjoint sub-batches} $\{\B_m\}_{m=1}^M$ of equal size. \label{alg:divide_step}
    \For{$m=1$ {\bfseries to} $M$ {\bfseries in parallel}} \label{alg:separated_grad_start}
    \State Draw a noise sample $\bxi_m \sim \bXi$.
    \State Compute the gradient $\g_m = \nabla_{\bo} \L(\bo; \B_m) \big|_{\bo^{(t)} \circ \bxi}$. \label{alg:separate_grad_end}
    \EndFor
    \State Compute the average gradient: $\g = \frac{1}{M} \sum_{m=1}^M \g_m$. \label{alg:avg_grad}
    \State Update the weights: $\bo^{(t+1)} = \bo^{(t)} - \eta_t \left(\g + \lambda \bo^{(t)}\right)$.
    \EndFor
 \end{algorithmic}
 \end{algorithm}
 

%% file: resnet50_imagenet_table.tex
\begin{table}[t]
\caption{{\bf DAMP surpasses the baselines on corrupted images in most cases and on average.} We report the predictive errors (lower is better) averaged over 3 seeds for the ResNet50 / ImageNet experiments. Subscript numbers represent standard deviations. We evaluate the models on IN-\{C, $\Cbar$, A, D, Sketch, Drawing, Cartoon, Hard\}, and adversarial examples generated by FGSM. For FGSM, we use $\epsilon=2/224$. For IN-\{C, $\Cbar$\}, we report the results averaged over all corruption types and severity levels. We use 90 epochs and the basic Inception-style preprocessing for all experiments.}
\label{tab:resnet50_imagenet}
\centering
\resizebox{\textwidth}{!}{%
\begin{tabular}{@{}cccccccccccc@{}}
\toprule
\multirowcell{2.5}{Method} & \multirowcell{2.5}{Clean\\Error (\%) $\downarrow$} & \multicolumn{10}{c}{Corrupted Error (\%) $\downarrow$}                                                                                                                                                                                                                                                                                                                                     \\ \cmidrule(l){3-12} 
                           &                                                    & FGSM                                  & A                                      & C                                     & $\Cbar$                                & Cartoon                                & D                                     & Drawing                               & Sketch                                & Hard                                  & Avg             \\ \midrule
Dropout                    & $23.6_{\color{darkgray}0.2}$                       & $90.7_{\color{darkgray}0.2}$          & $\mathbf{95.7_{\color{darkgray}<0.1}}$ & $61.7_{\color{darkgray}0.2}$          & $61.6_{\color{darkgray}<0.1}$          & $49.6_{\color{darkgray}0.2}$           & $88.9_{\color{darkgray}<0.1}$         & $77.4_{\color{darkgray}1.3}$          & $78.3_{\color{darkgray}0.3}$          & $85.8_{\color{darkgray}0.1}$          & $76.6$          \\
DAMP                       & $23.8_{\color{darkgray}<0.1}$                      & $\mathbf{88.3_{\color{darkgray}0.1}}$ & $96.2_{\color{darkgray}<0.1}$          & $\mathbf{58.6_{\color{darkgray}0.1}}$ & $\mathbf{58.7_{\color{darkgray}<0.1}}$ & $\mathbf{44.4_{\color{darkgray}<0.1}}$ & $88.7_{\color{darkgray}<0.1}$         & $\mathbf{71.1_{\color{darkgray}0.5}}$ & $\mathbf{76.3_{\color{darkgray}0.2}}$ & $85.3_{\color{darkgray}0.2}$          & $\mathbf{74.2}$ \\
SAM                        & $23.2_{\color{darkgray}<0.1}$                      & $90.4_{\color{darkgray}0.2}$          & $96.6_{\color{darkgray}0.1}$           & $60.2_{\color{darkgray}0.2}$          & $60.7_{\color{darkgray}0.1}$           & $47.6_{\color{darkgray}0.1}$           & $\mathbf{88.3_{\color{darkgray}0.1}}$ & $74.8_{\color{darkgray}<0.1}$         & $77.5_{\color{darkgray}0.1}$          & $85.8_{\color{darkgray}0.3}$          & $75.8$          \\
ASAM                       & $\mathbf{22.8_{\color{darkgray}0.1}}$              & $89.7_{\color{darkgray}0.2}$          & $96.8_{\color{darkgray}0.1}$           & $58.9_{\color{darkgray}0.1}$          & $59.2_{\color{darkgray}0.1}$           & $45.5_{\color{darkgray}<0.1}$          & $88.7_{\color{darkgray}0.1}$          & $72.3_{\color{darkgray}0.1}$          & $76.4_{\color{darkgray}0.2}$          & $\mathbf{85.2_{\color{darkgray}0.1}}$ & $74.7$          \\ \bottomrule
\end{tabular}%
}
\end{table}

%% file: vits16_imagenet_table.tex
\begin{table}[t]
\centering
\caption{{\bf ViT-S16 / ImageNet (IN) with basic Inception-style data augmentations}. Due to the high training cost, we report the predictive error (lower is better) of a single run. We evaluate corruption robustness of the models using IN-\{C, $\Cbar$, A, D, Sketch, Drawing, Cartoon, Hard\}, and adversarial examples generated by FGSM. For IN-\{C, $\Cbar$\}, we report the results averaged over all corruption types and severity levels. For FGSM, we use $\epsilon=2/224$. We also report the runtime of each experiment, showing that SAM and ASAM take twice as long to run than DAMP and AdamW given the same number of epochs. DAMP produces the most robust model on average.}
\label{tab:vits16_imagenet}
\resizebox{\textwidth}{!}{%
\begin{tabular}{@{}cccccccccccccc@{}}
\toprule
\multirowcell{2.5}{Method} & \multirowcell{2.5}{\#Epochs} & \multirowcell{2.5}{Runtime} & \multirowcell{2.5}{Clean\\Error (\%) $\downarrow$} & \multicolumn{10}{c}{Corrupted Error (\%) $\downarrow$}                                                                                                                  \\ \cmidrule(l){5-14} 
                           &                              &                             &                                                    & FGSM           & A              & C              & $\Cbar$        & Cartoon        & D              & Drawing        & Sketch         & Hard           & Avg            \\ \midrule
\multirowcell{2}{Dropout}  & 100                          & 20.6h                       & 28.55                                              & 93.47          & 93.44          & 65.87          & 64.52          & 50.37          & 91.15          & 79.62          & 88.06          & 87.19          & 79.30          \\
                           & 200                          & 41.1h                       & 28.74                                              & 90.95          & 93.33          & 66.90          & 64.83          & 51.23          & 92.56          & 81.24          & 87.99          & 87.60          & 79.63          \\ \midrule
\multirowcell{2}{DAMP}     & 100                          & 20.7h                       & 25.50                                              & 92.76          & 92.92          & 57.85          & 57.02          & 44.78          & 88.79          & 69.92          & 83.16          & 85.65          & 74.76          \\
                           & 200                          & 41.1h                       & \textbf{23.75}                                     & \textbf{84.33} & \textbf{90.56} & 55.58          & \textbf{55.58} & 41.06          & \textbf{87.87} & 68.36          & 81.82          & \textbf{84.18} & \textbf{72.15} \\ \midrule
SAM                        & 100                          & 41h                         & 23.91                                              & 87.61          & 93.96          & 55.56          & 55.93          & 42.53          & 88.23          & 69.53          & 81.86          & 85.54          & 73.42          \\
ASAM                       & 100                          & 41.1h                       & 24.01                                              & 85.85          & 92.99          & \textbf{55.13} & 55.64          & \textbf{40.74} & 89.03          & \textbf{67.80} & \textbf{81.47} & 84.31          & 72.55          \\ \bottomrule
\end{tabular}%
}
\end{table}

%% file: vit_imagenet_table_advanceaug.tex
\begin{table}[t]
\centering
\caption{{\bf ViT / ImageNet (IN) with MixUp and RandAugment}. We train ViT-S16 and ViT-B16 on ImageNet from scratch with advanced data augmentations (DAs). We evaluate the models on IN-\{C, $\Cbar$, A, D, Sketch, Drawing, Cartoon, Hard\}, and adversarial examples generated by FGSM. For FGSM, we use $\epsilon=2/224$. For IN-\{C, $\Cbar$\}, we report the results averaged over all corruption types and severity levels. These results indicate that: (i) DAMP can be combined with modern DA techniques to further enhance robustness; (ii) DAMP is capable of training large models like ViT-B16; (iii) given the same amount of training time, it is better to train a large model (ViT-B16) using DAMP than to train a smaller model (ViT-S16) using SAM/ASAM.}
\label{tab:vit_aug}
\resizebox{\textwidth}{!}{%
\begin{tabular}{@{}ccccccccccccccc@{}}
\toprule
\multirowcell{2.5}{Model}   & \multirowcell{2.5}{Method} & \multirowcell{2.5}{\#Epochs} & \multirowcell{2.5}{Runtime} & \multirowcell{2.5}{Clean\\Error (\%) $\downarrow$} & \multicolumn{10}{c}{Corrupted Error (\%) $\downarrow$}                                                                                                                  \\ \cmidrule(l){6-15} 
                            &                            &                              &                             &                                                    & FGSM           & $\Cbar$        & A              & C              & Cartoon        & D              & Drawing        & Sketch         & Hard           & Avg            \\ \midrule
\multirowcell{4}{ViT\\S16}  & Dropout                    & 500                          & 111h                        & 20.25                                              & 62.45          & 40.85          & 84.29          & 44.72          & 34.35          & 86.59          & 56.31          & 71.03          & 80.87          & 62.38          \\
                            & DAMP                       & 500                          & 111h                        & \textbf{20.09}                                     & 59.87          & \textbf{39.30} & \textbf{83.12} & \textbf{43.18} & 34.01          & \textbf{84.74} & \textbf{54.16} & \textbf{68.03} & \textbf{80.05} & \textbf{60.72} \\
                            & SAM                        & 300                          & 123h                        & 20.17                                              & 59.92          & 40.05          & 83.91          & 44.04          & 34.34          & 85.99          & 55.63          & 70.85          & 80.18          & 61.66          \\
                            & ASAM                       & 300                          & 123h                        & 20.38                                              & \textbf{59.38} & 39.44          & 83.64          & 43.41          & \textbf{33.82} & 85.41          & 54.43          & 69.13          & 80.50          & 61.02          \\ \midrule
\multirowcell{4}{ViT\\B16}  & Dropout                    & 275                          & 123h                        & 20.41                                              & 56.43          & 39.14          & 82.85          & 43.82          & 33.13          & 87.72          & 56.15          & 71.36          & 79.13          & 61.08          \\
                            & DAMP                       & 275                          & 124h                        & \textbf{19.36}                                     & \textbf{55.20} & 37.77          & \textbf{80.49} & 41.67          & 31.63          & \textbf{87.06} & 52.32          & \textbf{67.91} & \textbf{78.69} & \textbf{59.19} \\ 
                            & SAM                        & 150                          & 135h                        & 19.84                                              & 61.85          & 39.09          & 82.69          & 43.53          & 32.95          & 88.38          & 55.33          & 71.22          & 79.48          & 61.61 \\
                            & ASAM                       & 150                          & 136h                        & 19.40                                              & 58.87          & \textbf{37.41} & 82.21          & \textbf{41.18} & \textbf{30.76} & 88.03          & \textbf{51.84} & 69.54          & 78.83          & 59.85 \\
                            \bottomrule
\end{tabular}%
}
\end{table}    

%% file: appendix.tex
\section{Proof of Lemma~\ref{lemma:per_sample_loss_bound}}\label{sec:lemma_1_proof}
\input{lemma_1_proof.tex}
\section{Proof of Theorem~\ref{theorem:bounded_loss}}\label{sec:theorem_1_proof}
\input{theorem_1_proof.tex}
\section{Training with corruption}
\begin{figure}[h!]
    \centering
    \includegraphics[width=\textwidth]{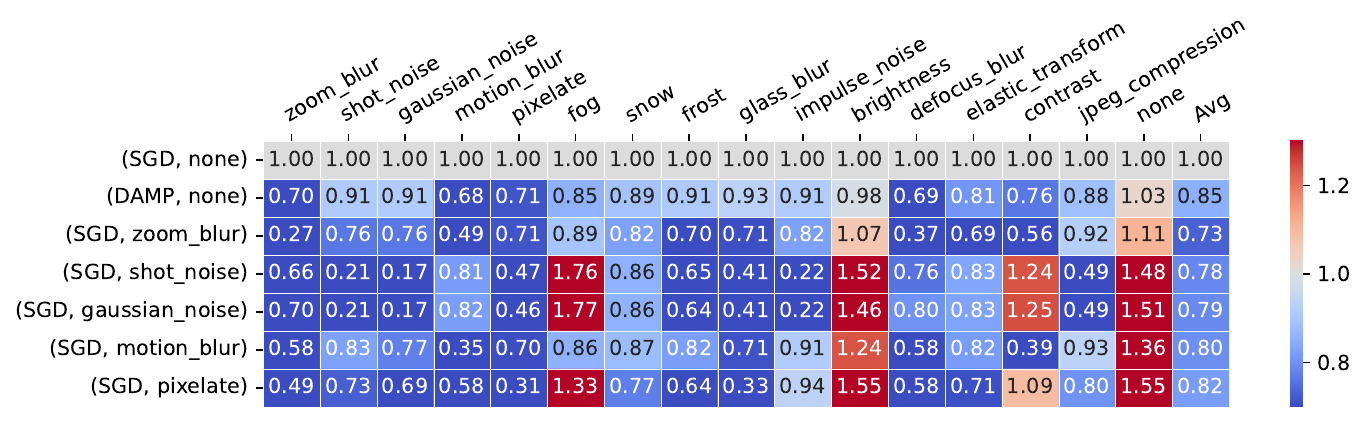}
    \caption{{\bf DAMP improves robustness to all corruptions while preserving accuracy on clean images.} Results of ResNet18/CIFAR-10 experiments averaged over 3 seeds. The heatmap shows $\CE^{f}_c$ described in \cref{eq:ce_c_f}, where each row corresponds to a tuple of of training (\texttt{method}, \texttt{corruption}), while each column corresponds to the test corruption. The \texttt{Avg} column shows the average of the results of the previous columns. \texttt{none} indicates no corruption. We use the models trained under the \texttt{SGD/none} setting (first row) as baselines to calculate the $\CE^{f}_c$. The last five rows are the 5 best training corruptions ranked by the results in the \texttt{Avg} column.}
    \label{fig:cifar10_corr_experiment_results}
\end{figure}

\begin{figure}[h!]
    \centering
    \includegraphics[width=\textwidth]{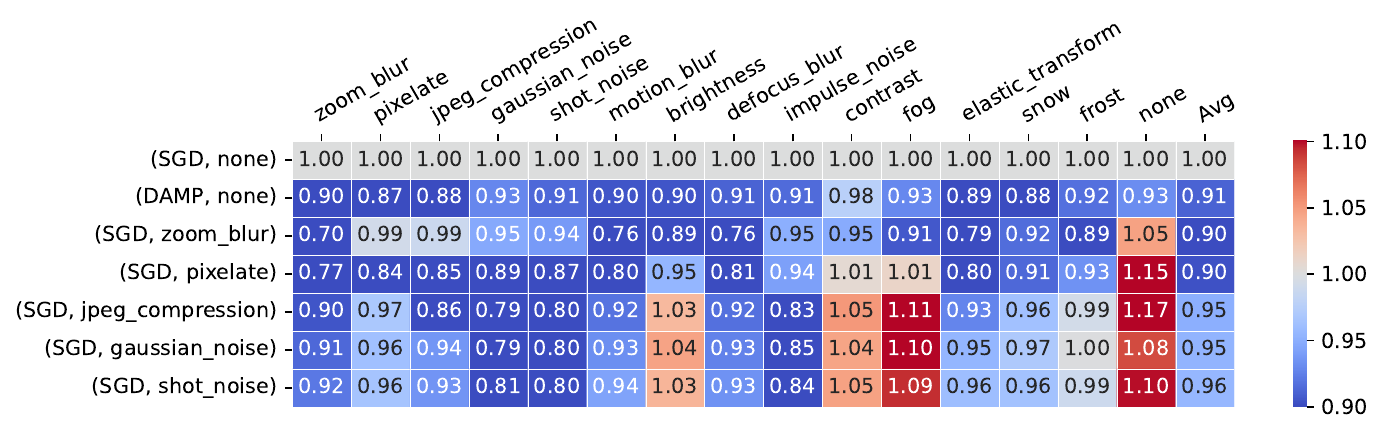}
    \caption{{\bf DAMP improves robustness to all corruptions while preserving accuracy on clean images.} Results of PreActResNet18/TinyImageNet experiments averaged over 3 seeds. The heatmap shows $\CE^{f}_c$ described in \cref{eq:ce_c_f}, where each row corresponds to a tuple of training (\texttt{method}, \texttt{corruption}), while each column corresponds to the test corruption. The \texttt{Avg} column shows the average of the results of the previous columns. \texttt{none} indicates no corruption. We use the models trained under the \texttt{SGD/none} setting (first row) as baselines to calculate the $\CE^{f}_c$. The last five rows are the 5 best training corruptions ranked by the results in the \texttt{Avg} column.}
    \label{fig:tinyimagenet_corr_experiment_results}
\end{figure}

Here we present \cref{alg:corr_training} which uses corruptions as data augmentation during training, as well as the experiment results of \cref{sec:damp_vs_corr} for ResNet18/CIFAR-10 and PreActResNet18/TinyImageNet settings in \cref{fig:cifar10_corr_experiment_results,fig:tinyimagenet_corr_experiment_results}.
\begin{algorithm}[h!]
    \caption{Training with corruption}\label{alg:corr_training}
 \begin{algorithmic}[1]
    \State {\bfseries Input:} training data $\S = \{(\x_k, y_k)\}_{k=1}^N$, a neural network $\f(\cdot; \bo)$ parameterized by $\bo \in \R^P$, number of iterations $T$, step sizes $\{\eta_t\}_{t=1}^T$, batch size $B$, a corruption $\g$ such as Gaussian noise, weight decay coefficient $\lambda$, a loss function $\L: \R^P \rightarrow \R_+$.
    \State {\bfseries Output:} Optimized parameter $\bo^{(T)}$.t
    \State Initialize parameter $\bo^{(0)}$.
    \For{$t=1$ {\bfseries to} $T$}
    \State Draw a mini-batch $\B=\{(\x_b, y_b)\}_{b=1}^B \sim \S$.
    \State Divide the mini-batch into two disjoint sub-batches of equal size $\B_1$ and $\B_2$.
    \State Apply the corruption $\g$ to all samples in $\B_1$: $\g(\B_1) = \{(\g(\x), y)\}_{(\x,y) \in \B_1}$.
    \State Compute the gradient $\g = \nabla_{\bo} \L(\bo; \g(\B_1) \cup \B_2)$.
    \State Update the weights: $\bo^{(t+1)} = \bo^{(t)} - \eta_t \left(\g + \lambda \bo^{(t)}\right)$.
    \EndFor
 \end{algorithmic}
 \end{algorithm}

\section{Training with random additive weight perturbations}
\begin{figure}[ht]
    \centering
    \includegraphics[width=.8\linewidth]{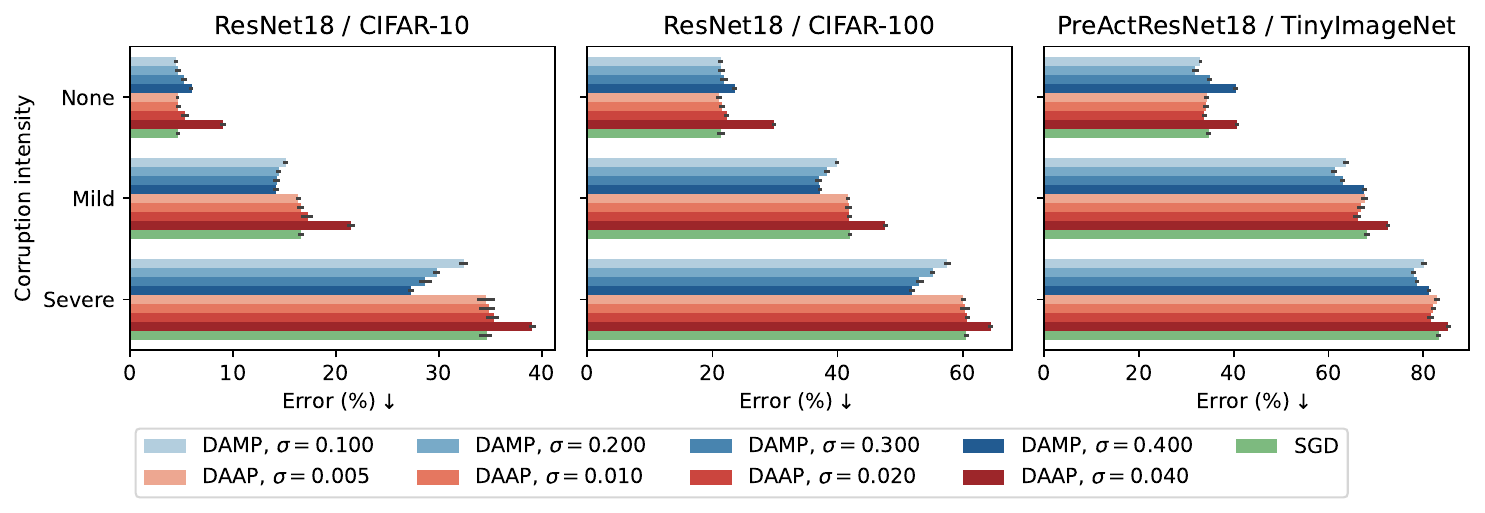}
    \caption{{\bf DAMP has better corruption robustness than DAAP.} We report the predictive errors (lower is better) averaged over 5 seeds. \texttt{None} indicates no corruption. \texttt{Mild} includes severity levels 1, 2 and 3. \texttt{Severe} includes severity levels 4 and 5. We evaluate DAMP and DAAP under different noise standard deviations $\sigma$. These results imply that the multiplicative weight perturbations of DAMP are more effective than the additive perturbations of DAAP in improving robustness to corruptions.}
    \label{fig:daap_vs_damp}
\end{figure}

Here, we present \cref{alg:daap} used in \cref{sec:damp_vs_daap} which trains DNNs under random additive weight perturbations and \cref{fig:daap_vs_damp} comparing performance between DAMP and DAAP.
\begin{algorithm}[ht]
\caption{DAAP: Data Augmentation via Additive Perturbations}\label{alg:daap}
\begin{algorithmic}[1]
\State {\bfseries Input:} training data $\S = \{(\x_k, y_k)\}_{k=1}^N$, a neural network $\f(\cdot; \bo)$ parameterized by $\bo \in \R^P$, number of iterations $T$, step sizes $\{\eta_t\}_{t=1}^T$, number of sub-batch $M$, batch size $B$ divisible by $M$, a noise distribution $\bXi = \N(\mathbf{0}, \sigma^2\mathbf{I}_P)$, weight decay coefficient $\lambda$, a loss function $\L: \R^P \rightarrow \R_+$.
\State {\bfseries Output:} Optimized parameter $\bo^{(T)}$.
\State Initialize parameter $\bo^{(0)}$.
\For{$t=1$ {\bfseries to} $T$}
\State Draw a mini-batch $\B=\{(\x_b, y_b)\}_{b=1}^B \sim \S$.
\State Divide the mini-batch into $M$ \emph{disjoint sub-batches} $\{\B_m\}_{m=1}^M$ of equal size.
\For{$m=1$ {\bfseries to} $M$ {\bfseries in parallel}}
\State Draw a noise sample $\bxi_m \sim \bXi$.
\State Compute the gradient $\g_m = \nabla_{\bo} \L(\bo; \B_m) \big|_{\bo^{(t)} + \bxi}$.
\EndFor
\State Compute the average gradient: $\g = \frac{1}{M} \sum_{m=1}^M \g_m$.
\State Update the weights: $\bo^{(t+1)} = \bo^{(t)} - \eta_t \left(\g + \lambda \bo^{(t)}\right)$.
\EndFor
\end{algorithmic}
\end{algorithm}
\section{Corruption datasets}\label{sec:corruption_datasets}
\paragraph{CIFAR-10/100-C \citep{hendrycks2019benchmarking}} These datasets contain the corrupted versions of the CIFAR-10/100 test sets. They contain 19 types of corruption, each divided into 5 levels of severity.
\paragraph{TinyImageNet-C \citep{hendrycks2019benchmarking}} This dataset contains the corrupted versions of the TinyImageNet test set. It contains 19 types of corruption, each divided into 5 levels of severity.
\paragraph{ImageNet-C \citep{hendrycks2019benchmarking}} This dataset contains the corrupted versions of the ImageNet validation set, as the labels of the true ImageNet test set was never released. It contains 15 types of corruption, each divided into 5 levels of severity.
\paragraph{ImageNet-$\Cbar$ \citep{mintun2021on}} This dataset contains the corrupted versions of the ImageNet validation set, as the labels of the true ImageNet test set was never released. It contains 10 types of corruption, each divided into 5 levels of severity. The types of corruption in ImageNet-$\Cbar$ differ from those in ImageNet-C.
\paragraph{ImageNet-A \citep{imagenet-a}} This dataset contains natural adversarial examples, which are real-world, unmodified, and naturally occurring examples that cause machine learning model performance to significantly degrade.
The images contain in this dataset, while differ from those in the ImageNet validation set, stills belong to the same set of classes.
\paragraph{ImageNet-D \citep{imagenet-d}} This dataset contains images belong to the classes of ImageNet but they are modified by diffusion models to change the background, material, and texture.
\paragraph{ImageNet-Cartoon and ImageNet-Drawing \citep{imagenet-cartoon-drawing}} This dataset contains the drawing and cartoon versions of the images in the ImageNet validation set.
\paragraph{ImageNet-Sketch \citep{imagenet-sketch}} This dataset contains sketch images belonging to the classes of the ImageNet dataset.
\paragraph{ImageNet-Hard \citep{taesiri2023zoom}} This dataset comprises an array of challenging images, curated from several validation datasets of ImageNet.

\section{Training details}\label{sec:training_detail}
For each method and each setting, we tune the important hyperparameters ($\sigma$ for DAMP, $\rho$ for SAM and ASAM) using $10\%$ of the training set as validation set.
\paragraph{CIFAR-10/100} For each setting, we train a ResNet18 for 300 epochs. We use a batch size of 128. We use a learning rate of $0.1$ and a weight decay coefficient of $5 \times 10^{-4}$. We use SGD with Nesterov momentum as the optimizer with a momentum coefficient of $0.9$. The learning rate is kept at $0.1$ until epoch 150, then is linearly annealed to $0.001$ from epoch 150 to epoch 270, then kept at $0.001$ for the rest of the training. We use basic data preprocessing, which includes channel-wise normalization, random cropping after padding and random horizontal flipping. On CIFAR-10, we set $\sigma = 0.2$ for DAMP, $\rho=0.045$ for SAM and $\rho=1.0$ for ASAM. On CIFAR-100, we set $\sigma=0.1$ for DAMP, $\rho=0.06$ for SAM and $\rho=2.0$ for ASAM. Each method is trained on a single host with 8 Nvidia V100 GPUs where the data batch is evenly distributed among the GPUs at each iteration (data parallelism). This means we use the number of sub-batches $M = 8$ for DAMP.
\paragraph{TinyImageNet} For each setting, we train a PreActResNet18 for 150 epochs. We use a batch size of 128. We use a learning rate of $0.1$ and a weight decay coefficient of $2.5 \times 10^{-4}$. We use SGD with Nesterov momentum as the optimizer with a momentum coefficient of $0.9$. The learning rate is kept at $0.1$ until epoch 75, then is linearly annealed to $0.001$ from epoch 75 to epoch 135, then kept at $0.001$ for the rest of the training. We use basic data preprocessing, which includes channel-wise normalization, random cropping after padding and random horizontal flipping. We set $\sigma=0.2$ for DAMP, $\rho=0.2$ for SAM and $\rho=3.0$ for ASAM. Each method is trained on a single host with 8 Nvidia V100 GPUs where the data batch is evenly distributed among the GPUs at each iteration (data parallelism). This means we use the number of sub-batches $M = 8$ for DAMP.
\paragraph{ResNet50 / ImageNet} We train each experiment for 90 epochs. We use a batch size of 2048. We use a weight decay coefficient of $1 \times 10^{-4}$. We use SGD with Nesterov momentum as the optimizer with a momentum coefficient of $0.9$. We use basic Inception-style data preprocessing, which includes random cropping, resizing to the resolution of $224 \times 224$, random horizontal flipping and channel-wise normalization. We increase the learning rate linearly from $8 \times 10^{-4}$ to $0.8$ for the first 5 epochs then decrease the learning rate from $0.8$ to $8 \times 10^{-4}$ using a cosine schedule for the remaining epochs. All experiments were run on a single host with 8 Nvidia V100 GPUs and we set $M=8$ for DAMP. We use $p=0.05$ for Dropout, $\sigma=0.1$ for DAMP, $\rho=0.05$ for SAM, and $\rho=1.5$ for ASAM. We also use the image resolution of $224 \times 224$ during evaluation.
\paragraph{ViT-S16 / ImageNet / Basic augmentations} We follow the training setup of \citet{beyer2022better} with one difference is that we only use basic Inception-style data processing similar to the ResNet50/ImageNet experiments. We use AdamW as the optimizer with $\beta_1=0.9$, $\beta_2=0.999$ and $\epsilon=10^{-8}$. We clip the gradient norm to $1.0$. We use a weight decay coefficient of $0.1$. We use a batch size of $1024$. We increase the learning rate linearly from $10^{-6}$ to $10^{-3}$ for the first 10000 iterations, then we anneal the learning rate from $10^{-3}$ to $0$ using a cosine schedule for the remaining iterations. We use the image resolution of $224 \times 224$ for both training and testing. Following \citet{beyer2022better}, we make 2 minor modifications to the original ViT-S16 architecture: (1) We change the position embedding layer from \texttt{learnable} to \texttt{sincos2d}; (2) We change the input of the final classification layer from the embedding of the \texttt{[cls]} token to global average-pooling. All experiments were run on a single host with 8 Nvidia V100 GPUs and we set $M=8$ for DAMP. We use $p=0.10$ for Dropout, $\sigma=0.25$ for DAMP, $\rho=0.6$ for SAM, and $\rho=3.0$ for ASAM.

\paragraph{ViT-S16 and B16 / ImageNet / MixUp and RandAugment} Most of the hyperparameters are identical to the ViT-S16 / ImageNet / Basic augmentations setting. With ViT-S16, we use $p=0.1$ for Dropout, $\sigma=0.10$ for DAMP, $\rho=0.015$ for SAM, and $\rho=0.4$ for ASAM. With ViT-B16, we use $p=0.1$ for Dropout, $\sigma=0.15$ for DAMP, $\rho=0.025$ for SAM, and $\rho=0.6$ for ASAM. 

%% file: lemma_1_proof.tex
\begin{proof}
   Here we note that:
   \begin{align}
      \x^{(h)} &\defeq \f^{(h)}(\x) \\
      \x_\g^{(h)} &\defeq \f^{(h)}(\g(\x)) \\
      \bd_\g \ell(\bo, \x, y) &\defeq \ell(\bo, \x_\g, y) - \ell(\bo, \x, y) \\
      \bd_\g \x^{(h)}     &\defeq \x^{(h)}_\g - \x^{(h)}
   \end{align}
   We first notice that the per-sample loss $\ell(\bo,\x,y)$ can be viewed as a function of the intermediate activation $\x^{(h)}$ of layer $h$ (see \cref{fig:fig2}). From \cref{assumption:lipschitz_grad}, there exists a constant $L_h > 0$ such that:
   \begin{equation}
      \|\nabla_{\x^{(h)}_\g} \ell(\bo,\x_\g,y) - \nabla_{\x^{(h)}} \ell(\bo,\x,y)\|_2 \leq L_h\|\bd_\g \x^{(h)}\|_2
   \end{equation}
   which gives us the following quadratic bound:
   \begin{align}\label{eq:quadratic_bound}
      \ell(\bo, \x_\g, y) \leq \ell(\bo, \x, y) + \left\langle \nabla_{\x^{(h)}} \ell(\bo,\x,y), \bd_\g \x^{(h)} \right\rangle + \frac{L_h}{2}\|\bd_\g \x^{(h)}\|_2^2
   \end{align}
   where $\langle \cdot, \cdot \rangle$ denotes the dot product between two vectors. The results in the equation above have been proven in \citet{bohning1988monotonicity}. Subtracting $\ell(\bo, \x, y)$ from both side of \cref{eq:quadratic_bound} gives us:
   \begin{equation}
      \bd_\g \ell(\bo, \x, y) \leq \left\langle \nabla_{\x^{(h)}} \ell(\bo,\x,y), \bd_\g \x^{(h)} \right\rangle + \frac{L_h}{2}\|\bd_\g \x^{(h)}\|_2^2
   \end{equation}
   Since the pre-activation output of layer $h+1$ is $\z^{(h+1)}(\x) = \W^{(h+1)} \f^{(h)}(\x) = \W^{(h+1)} \x^{(h)}$, we can rewrite the inequality above as:
   \begin{equation}\label{eq:lemma_1_step1}
      \bd_\g \ell(\bo, \x, y) \leq {\left\langle \nabla_{\z^{(h+1)}} \ell(\bo, \x, y) \otimes \bd_\g\x^{(h)}, \W^{(h+1)} \right\rangle}_F + \frac{L_h}{2}\|\bd_\g \x^{(h)}\|_2^2
   \end{equation}
   where $\otimes$ denotes the outer product of two vectors and $\langle \cdot, \cdot \rangle_F$ denotes the Frobenius inner product of two matrices of similar dimension.

   From \cref{assumption:bounded_g}, we have that there exists a constant $M > 0$ such that:
   \begin{equation}
      \|\bd_\g \x^{(0)}\|_2^2 = \|\x^{(0)}_\g - \x^{(0)}\|_2^2 = \|\g(\x) - \x\|_2^2 \leq M
   \end{equation}
   Given that $\x^{(1)} = \bs^{(1)}\left(\W^{(1)}\x^{(0)}\right)$, we have:
   \begin{align}
      \|\bd_\g \x^{(1)}\|_2^2 = \|\x^{(1)}_\g - \x^{(1)}\|_2^2 \leq \|\W^{(1)}\bd_\g \x^{(0)}\|_2^2 
   \end{align}
   Here we assume that the activate $\bs$ satisfies $\|\bs(\x) - \bs(\y)\|_2 \leq \|\x-\y\|_2$, which is true for modern activation functions such as ReLU.
   Since $\|\bd_\g \x^{(0)}\|_2^2$ is bounded, there exists a constant $\hat{C}_{\g}^{(1)}(\x)$ such that:
   \begin{equation}
    \|\bd_\g \x^{(1)}\|_2^2 = \|\x^{(1)}_\g - \x^{(1)}\|_2^2 \leq \|\W^{(1)}\bd_\g \x^{(0)}\|_2^2 \leq \frac{\hat{C}_{\g}^{(1)}(\x)}{2} \|\W^{(1)}\|_F^2
   \end{equation}
   where $\|\cdot\|_F$ denotes the Frobenius norm. Similarly, as we have proven that $\|\bd_\g \x^{(1)}\|_2^2$ is bounded, there exists a constant $\hat{C}_{\g}^{(2)}(\x)$ such that:
   \begin{equation}
    \|\bd_\g \x^{(2)}\|_2^2 = \|\x^{(2)}_\g - \x^{(2)}\|_2^2 \leq \|\W^{(2)}\bd_\g \x^{(1)}\|_2^2 \leq \frac{\hat{C}_{\g}^{(2)}(\x)}{2} \|\W^{(2)}\|_F^2
   \end{equation}
   Thus we have proven that for all $h=1,\dots,H$, there exists a constant $\hat{C}_{\g}^{(h)}(\x)$ such that:
   \begin{equation}\label{eq:lemma_1_step2}
      \|\bd_\g \x^{(h)}\|_2^2 \leq \frac{\hat{C}_{\g}^{(h)}(\x)}{2} \|\W^{(h)}\|_F^2
   \end{equation}
   By combining \cref{eq:lemma_1_step1,eq:lemma_1_step2} and setting $C^{(h)}_\g(\x)=L_h \hat{C}_{\g}^{(h)}(\x)$, we arrive at \cref{eq:lemma_1}.
\end{proof}

%% file: theorem_1_proof.tex
\begin{proof} From \cref{lemma:per_sample_loss_bound}, we have for all $h=0,\dots,H-1$:
    \begin{align}
        &\L(\bo; \g(\S)) = \frac{1}{N} \sum_{k=1}^N \ell(\bo, \g(\x_k), y_k)
         = \frac{1}{N} \sum_{k=1}^N \bigg(\ell(\bo, \x_k, y_k) + \bd_\g \ell(\bo, \x_k, y_k)\bigg) \\
        & \leq \L(\bo; \S) + \frac{1}{N}\sum_{k=1}^N {\left\langle \nabla_{\z^{(h+1)}} \ell(\bo,\x_k,y_k) \otimes \bd_\g\x^{(h)}_k, \W^{(h+1)} \right\rangle}_F + \frac{\hat{C}_\g^{(h)}}{2} \|\W^{(h)}\|_F^2 \label{eq:per_layer_training_loss_bound}
    \end{align}
    where $\hat{C}_\g^{(h)} = \max_{\x \in \S} C_\g^{(h)}(\x)$. Since this bound is true for all $h$, we can take the average:
    \begin{multline}
        \L(\bo; \g(\S)) \leq \L(\bo; \S) + \frac{1}{H}\sum_{h=1}^H \frac{1}{N}\sum_{k=1}^N {\left\langle \nabla_{\z^{(h)}} \ell(\bo,\x_k,y_k) \otimes \bd_\g\x^{(h-1)}_k, \W^{(h)} \right\rangle}_F \\
        + \frac{C_\g}{2} \|\bo\|_F^2 \label{eq:avg_bound}
    \end{multline}
    where $C_\g = \frac{1}{H}\sum_{h=1}^H \hat{C}_{\g}^{(h)}$. The right-hand side of \cref{eq:avg_bound} can be written as:
    % \cref{eq:avg_training_loss_shift} can be written as:
    \begin{align}
        &\L(\bo; \S) + \frac{1}{H}\sum_{h=1}^H {\left\langle \frac{1}{N}\sum_{k=1}^N \nabla_{\z^{(h)}} \ell(\bo,\x_k,y_k) \otimes \bd_\g\x^{(h-1)}_k, \W^{(h)} \right\rangle}_F + \frac{C_\g}{2} \|\bo\|_F^2\\
        &= \L(\bo; \S) + \sum_{h=1}^H {\left\langle \nabla_{\W^{(h)}} \L(\bo;\S), \W^{(h)} \circ \bxi^{(h)}(\g)\right\rangle}_F + \frac{C_\g}{2} \|\bo\|_F^2 \\
        &\leq \L(\bo + \bo \circ \bxi(\g); \S) + \frac{C_\g}{2} \|\bo\|_F^2 = \L(\bo \circ \left(1 + \bxi(\g)\right); \S) + \frac{C_\g}{2} \|\bo\|_F^2\label{eq:bound_in_weight_space}
    \end{align}
    where $\bxi^{(h)}(\g)$ is a matrix of the same dimension as $\W^{(h)}$ whose each entry is defined as:
    \begin{equation}
        \left[\bxi^{(h)}(\g)\right]_{i,j} = \frac{1}{H}\frac{\left[\sum_{k=1}^N \nabla_{\z^{(h)}}\ell(\bo,\x_k,y_k) \otimes \bd_\g\x^{(h-1)}_k\right]_{i,j}}{\left[\sum_{k=1}^N \nabla_{\z^{(h)}}\ell(\bo,\x_k,y_k) \otimes \x^{(h-1)}_k\right]_{i,j}}
    \end{equation}
    The inequality in \cref{eq:bound_in_weight_space} is due to the first-order Taylor expansion and the assumption that the training loss is locally convex at $\bo$. This assumption is expected to hold for the final solution but does not necessarily hold for any $\bo$. \cref{eq:bounded_loss} is obtained by combining \cref{eq:avg_bound} and \cref{eq:bound_in_weight_space}.
    \end{proof}

%% file: checklist.tex
\section*{NeurIPS Paper Checklist}
\begin{enumerate}

\item {\bf Claims}
    \item[] Question: Do the main claims made in the abstract and introduction accurately reflect the paper's contributions and scope?
    \item[] Answer: \answerYes{} % Replace by \answerYes{}, \answerNo{}, or \answerNA{}.
    \item[] Justification: We clearly state our contributions and claims in the abstract and introduction and back these claims and contributions with theoretical justifications and experimental results.

\item {\bf Limitations}
    \item[] Question: Does the paper discuss the limitations of the work performed by the authors?
    \item[] Answer: \answerYes{} % Replace by \answerYes{}, \answerNo{}, or \answerNA{}.
    \item[] Justification: We discuss the limitations of our work in \cref{sec:conclusion}, which outlines the shortcomings of our theoretical proofs and our experiments.

\item {\bf Theory Assumptions and Proofs}
    \item[] Question: For each theoretical result, does the paper provide the full set of assumptions and a complete (and correct) proof?
    \item[] Answer: \answerYes{} % Replace by \answerYes{}, \answerNo{}, or \answerNA{}.
    \item[] Justification: We provide full set of assumptions and complete proofs of the theoretical result in \cref{sec:damp} and \cref{sec:lemma_1_proof}.

    \item {\bf Experimental Result Reproducibility}
    \item[] Question: Does the paper fully disclose all the information needed to reproduce the main experimental results of the paper to the extent that it affects the main claims and/or conclusions of the paper (regardless of whether the code and data are provided or not)?
    \item[] Answer: \answerYes{} % Replace by \answerYes{}, \answerNo{}, or \answerNA{}.
    \item[] Justification: We provide all the details regarding our experiments in \cref{sec:training_detail}. We also describe the new algorithm that we propose in detail in \cref{alg:damp}.

\item {\bf Open access to data and code}
    \item[] Question: Does the paper provide open access to the data and code, with sufficient instructions to faithfully reproduce the main experimental results, as described in supplemental material?
    \item[] Answer: \answerYes{} % Replace by \answerYes{}, \answerNo{}, or \answerNA{}.
    \item[] Justification: We provide our code on a public GitHub repository with instruction on how to reproduce the experimental results.

\item {\bf Experimental Setting/Details}
    \item[] Question: Does the paper specify all the training and test details (e.g., data splits, hyperparameters, how they were chosen, type of optimizer, etc.) necessary to understand the results?
    \item[] Answer: \answerYes{} % Replace by \answerYes{}, \answerNo{}, or \answerNA{}.
    \item[] Justification: We provide all the details regarding our experiments in \cref{sec:training_detail} and specify sufficient details to understand the results in the main paper.

\item {\bf Experiment Statistical Significance}
    \item[] Question: Does the paper report error bars suitably and correctly defined or other appropriate information about the statistical significance of the experiments?
    \item[] Answer: \answerYes{} % Replace by \answerYes{}, \answerNo{}, or \answerNA{}.
    \item[] Justification: For \cref{fig:daap_vs_damp,fig:small_benchmark}, the error bars display $95\%$ confidence intervals. For \cref{tab:resnet50_imagenet}, we report the standard deviation. For \cref{fig:cifar100_corr_experiment_results,fig:cifar10_corr_experiment_results,fig:tinyimagenet_corr_experiment_results}, we cannot display the error bars since these figures show heatmaps. For \cref{tab:vits16_imagenet,tab:vit_aug}, we does not report the error bars since we ran each experiment once due to the high training cost and our limit in computing resources.

\item {\bf Experiments Compute Resources}
    \item[] Question: For each experiment, does the paper provide sufficient information on the computer resources (type of compute workers, memory, time of execution) needed to reproduce the experiments?
    \item[] Answer: \answerYes{} % Replace by \answerYes{}, \answerNo{}, or \answerNA{}.
    \item[] Justification: We state clearly in the paper that we run all experiments on a single machine with 8 Nvidia V100 GPUs.
    
\item {\bf Code Of Ethics}
    \item[] Question: Does the research conducted in the paper conform, in every respect, with the NeurIPS Code of Ethics \url{https://neurips.cc/public/EthicsGuidelines}?
    \item[] Answer: \answerYes{} % Replace by \answerYes{}, \answerNo{}, or \answerNA{}.
    \item[] Justification: We have read the NeurIPS Code of Ethics and confirm that the research in this paper conforms with these guidelines.

\item {\bf Broader Impacts}
    \item[] Question: Does the paper discuss both potential positive societal impacts and negative societal impacts of the work performed?
    \item[] Answer: \answerYes{} % Replace by \answerYes{}, \answerNo{}, or \answerNA{}.
    \item[] Justification: We discuss at the end of the paper the possible societal impact of the work in this paper.
    
\item {\bf Safeguards}
    \item[] Question: Does the paper describe safeguards that have been put in place for responsible release of data or models that have a high risk for misuse (e.g., pretrained language models, image generators, or scraped datasets)?
    \item[] Answer: \answerNA{} % Replace by \answerYes{}, \answerNo{}, or \answerNA{}.
    \item[] Justification: This paper does not release any new datasets or models.

\item {\bf Licenses for existing assets}
    \item[] Question: Are the creators or original owners of assets (e.g., code, data, models), used in the paper, properly credited and are the license and terms of use explicitly mentioned and properly respected?
    \item[] Answer: \answerYes{} % Replace by \answerYes{}, \answerNo{}, or \answerNA{}.
    \item[] Justification: We properly cite all the papers that produced the datasets and model architectures used in this work.

\item {\bf New Assets}
    \item[] Question: Are new assets introduced in the paper well documented and is the documentation provided alongside the assets?
    \item[] Answer: \answerYes{} % Replace by \answerYes{}, \answerNo{}, or \answerNA{}.
    \item[] Justification: We provide our code in a public GitHub repository with documentation on how to run the code.

\item {\bf Crowdsourcing and Research with Human Subjects}
    \item[] Question: For crowdsourcing experiments and research with human subjects, does the paper include the full text of instructions given to participants and screenshots, if applicable, as well as details about compensation (if any)? 
    \item[] Answer: \answerNA{} % Replace by \answerYes{}, \answerNo{}, or \answerNA{}.
    \item[] Justification: This work does not involve crowdsourcing nor human subjects.

\item {\bf Institutional Review Board (IRB) Approvals or Equivalent for Research with Human Subjects}
    \item[] Question: Does the paper describe potential risks incurred by study participants, whether such risks were disclosed to the subjects, and whether Institutional Review Board (IRB) approvals (or an equivalent approval/review based on the requirements of your country or institution) were obtained?
    \item[] Answer: \answerNA{} % Replace by \answerYes{}, \answerNo{}, or \answerNA{}.
    \item[] Justification: This work does not involve crowdsourcing nor human subjects.

\end{enumerate}

%% file: main.bbl
\begin{thebibliography}{49}
\providecommand{\natexlab}[1]{#1}
\providecommand{\url}[1]{\texttt{#1}}
\expandafter\ifx\csname urlstyle\endcsname\relax
  \providecommand{\doi}[1]{doi: #1}\else
  \providecommand{\doi}{doi: \begingroup \urlstyle{rm}\Url}\fi

\bibitem[Hendrycks and Dietterich(2019)]{hendrycks2019benchmarking}
Dan Hendrycks and Thomas Dietterich.
\newblock Benchmarking neural network robustness to common corruptions and
  perturbations.
\newblock \emph{arXiv preprint arXiv:1903.12261}, 2019.

\bibitem[Amodei et~al.(2016)Amodei, Olah, Steinhardt, Christiano, Schulman, and
  Man{\'e}]{amodei2016concrete}
Dario Amodei, Chris Olah, Jacob Steinhardt, Paul Christiano, John Schulman, and
  Dan Man{\'e}.
\newblock Concrete problems in ai safety.
\newblock \emph{arXiv preprint arXiv:1606.06565}, 2016.

\bibitem[Geirhos et~al.(2018)Geirhos, Temme, Rauber, Sch{\"u}tt, Bethge, and
  Wichmann]{geirhos2018generalisation}
Robert Geirhos, Carlos~RM Temme, Jonas Rauber, Heiko~H Sch{\"u}tt, Matthias
  Bethge, and Felix~A Wichmann.
\newblock Generalisation in humans and deep neural networks.
\newblock \emph{Advances in neural information processing systems}, 31, 2018.

\bibitem[Cubuk et~al.(2018)Cubuk, Zoph, Mane, Vasudevan, and
  Le]{cubuk2018autoaugment}
Ekin~D Cubuk, Barret Zoph, Dandelion Mane, Vijay Vasudevan, and Quoc~V Le.
\newblock Autoaugment: Learning augmentation policies from data.
\newblock \emph{arXiv preprint arXiv:1805.09501}, 2018.

\bibitem[Hendrycks et~al.(2019)Hendrycks, Mu, Cubuk, Zoph, Gilmer, and
  Lakshminarayanan]{hendrycks2019augmix}
Dan Hendrycks, Norman Mu, Ekin~D Cubuk, Barret Zoph, Justin Gilmer, and Balaji
  Lakshminarayanan.
\newblock Augmix: A simple data processing method to improve robustness and
  uncertainty.
\newblock \emph{arXiv preprint arXiv:1912.02781}, 2019.

\bibitem[Lopes et~al.(2019)Lopes, Yin, Poole, Gilmer, and
  Cubuk]{lopes2019improving}
Raphael~Gontijo Lopes, Dong Yin, Ben Poole, Justin Gilmer, and Ekin~D Cubuk.
\newblock Improving robustness without sacrificing accuracy with patch gaussian
  augmentation.
\newblock \emph{arXiv preprint arXiv:1906.02611}, 2019.

\bibitem[Mintun et~al.(2021)Mintun, Kirillov, and Xie]{mintun2021on}
Eric Mintun, Alexander Kirillov, and Saining Xie.
\newblock On interaction between augmentations and corruptions in natural
  corruption robustness.
\newblock In A.~Beygelzimer, Y.~Dauphin, P.~Liang, and J.~Wortman Vaughan,
  editors, \emph{Advances in Neural Information Processing Systems}, 2021.
\newblock URL \url{https://openreview.net/forum?id=LOHyqjfyra}.

\bibitem[Lakshminarayanan et~al.(2017)Lakshminarayanan, Pritzel, and
  Blundell]{lakshminarayanan2017simple}
Balaji Lakshminarayanan, Alexander Pritzel, and Charles Blundell.
\newblock Simple and scalable predictive uncertainty estimation using deep
  ensembles.
\newblock \emph{Advances in neural information processing systems}, 30, 2017.

\bibitem[Ovadia et~al.(2019)Ovadia, Fertig, Ren, Nado, Sculley, Nowozin,
  Dillon, Lakshminarayanan, and Snoek]{ovadia2019canyou}
Yaniv Ovadia, Emily Fertig, Jie Ren, Zachary Nado, D.~Sculley, Sebastian
  Nowozin, Joshua Dillon, Balaji Lakshminarayanan, and Jasper Snoek.
\newblock Can you trust your model\textquotesingle s uncertainty? {E}valuating
  predictive uncertainty under dataset shift.
\newblock In H.~Wallach, H.~Larochelle, A.~Beygelzimer, F.~d\textquotesingle
  Alch\'{e}-Buc, E.~Fox, and R.~Garnett, editors, \emph{Advances in Neural
  Information Processing Systems}, volume~32. Curran Associates, Inc., 2019.
\newblock URL
  \url{https://proceedings.neurips.cc/paper_files/paper/2019/file/8558cb408c1d76621371888657d2eb1d-Paper.pdf}.

\bibitem[Dusenberry et~al.(2020)Dusenberry, Jerfel, Wen, Ma, Snoek, Heller,
  Lakshminarayanan, and Tran]{dusenberry20a}
Michael Dusenberry, Ghassen Jerfel, Yeming Wen, Yian Ma, Jasper Snoek,
  Katherine Heller, Balaji Lakshminarayanan, and Dustin Tran.
\newblock Efficient and scalable {B}ayesian neural nets with rank-1 factors.
\newblock In Hal~Daumé III and Aarti Singh, editors, \emph{Proceedings of the
  37th International Conference on Machine Learning}, volume 119 of
  \emph{Proceedings of Machine Learning Research}, pages 2782--2792. PMLR,
  13--18 Jul 2020.
\newblock URL \url{https://proceedings.mlr.press/v119/dusenberry20a.html}.

\bibitem[Trinh et~al.(2022)Trinh, Heinonen, Acerbi, and Kaski]{trinh22a}
Trung Trinh, Markus Heinonen, Luigi Acerbi, and Samuel Kaski.
\newblock Tackling covariate shift with node-based {B}ayesian neural networks.
\newblock In Kamalika Chaudhuri, Stefanie Jegelka, Le~Song, Csaba Szepesvari,
  Gang Niu, and Sivan Sabato, editors, \emph{Proceedings of the 39th
  International Conference on Machine Learning}, volume 162 of
  \emph{Proceedings of Machine Learning Research}, pages 21751--21775. PMLR,
  17--23 Jul 2022.

\bibitem[Kwon et~al.(2021)Kwon, Kim, Park, and Choi]{kwon2021asam}
Jungmin Kwon, Jeongseop Kim, Hyunseo Park, and In~Kwon Choi.
\newblock Asam: Adaptive sharpness-aware minimization for scale-invariant
  learning of deep neural networks.
\newblock In \emph{International Conference on Machine Learning}, pages
  5905--5914. PMLR, 2021.

\bibitem[Dosovitskiy et~al.(2021)Dosovitskiy, Beyer, Kolesnikov, Weissenborn,
  Zhai, Unterthiner, Dehghani, Minderer, Heigold, Gelly, Uszkoreit, and
  Houlsby]{dosovitskiy2021an}
Alexey Dosovitskiy, Lucas Beyer, Alexander Kolesnikov, Dirk Weissenborn,
  Xiaohua Zhai, Thomas Unterthiner, Mostafa Dehghani, Matthias Minderer, Georg
  Heigold, Sylvain Gelly, Jakob Uszkoreit, and Neil Houlsby.
\newblock An image is worth 16x16 words: Transformers for image recognition at
  scale.
\newblock In \emph{International Conference on Learning Representations}, 2021.
\newblock URL \url{https://openreview.net/forum?id=YicbFdNTTy}.

\bibitem[He et~al.(2016{\natexlab{a}})He, Zhang, Ren, and Sun]{he2016deep}
Kaiming He, Xiangyu Zhang, Shaoqing Ren, and Jian Sun.
\newblock Deep residual learning for image recognition.
\newblock In \emph{IEEE conference on Computer Vision and Pattern Recognition},
  2016{\natexlab{a}}.

\bibitem[Szegedy et~al.(2016)Szegedy, Vanhoucke, Ioffe, Shlens, and
  Wojna]{szegedy2016rethinking}
Christian Szegedy, Vincent Vanhoucke, Sergey Ioffe, Jon Shlens, and Zbigniew
  Wojna.
\newblock Rethinking the inception architecture for computer vision.
\newblock In \emph{Proceedings of the IEEE conference on computer vision and
  pattern recognition}, pages 2818--2826, 2016.

\bibitem[Chen et~al.(2022)Chen, Hsieh, and Gong]{chen2022when}
Xiangning Chen, Cho-Jui Hsieh, and Boqing Gong.
\newblock When vision transformers outperform resnets without pre-training or
  strong data augmentations.
\newblock In \emph{International Conference on Learning Representations}, 2022.
\newblock URL \url{https://openreview.net/forum?id=LtKcMgGOeLt}.

\bibitem[Beyer et~al.(2022)Beyer, Zhai, and Kolesnikov]{beyer2022better}
Lucas Beyer, Xiaohua Zhai, and Alexander Kolesnikov.
\newblock Better plain vit baselines for imagenet-1k.
\newblock \emph{arXiv preprint arXiv:2205.01580}, 2022.

\bibitem[Zhang et~al.(2018)Zhang, Cisse, Dauphin, and
  Lopez-Paz]{zhang2018mixup}
Hongyi Zhang, Moustapha Cisse, Yann~N. Dauphin, and David Lopez-Paz.
\newblock mixup: Beyond empirical risk minimization.
\newblock \emph{International Conference on Learning Representations}, 2018.
\newblock URL \url{https://openreview.net/forum?id=r1Ddp1-Rb}.

\bibitem[Cubuk et~al.(2020)Cubuk, Zoph, Shlens, and Le]{cubuk2020randaugment}
Ekin~D Cubuk, Barret Zoph, Jonathon Shlens, and Quoc~V Le.
\newblock Randaugment: Practical automated data augmentation with a reduced
  search space.
\newblock In \emph{Proceedings of the IEEE/CVF conference on computer vision
  and pattern recognition workshops}, pages 702--703, 2020.

\bibitem[Rusak et~al.(2020)Rusak, Schott, Zimmermann, Bitterwolf, Bringmann,
  Bethge, and Brendel]{rusak2020simple}
Evgenia Rusak, Lukas Schott, Roland~S Zimmermann, Julian Bitterwolf, Oliver
  Bringmann, Matthias Bethge, and Wieland Brendel.
\newblock A simple way to make neural networks robust against diverse image
  corruptions.
\newblock \emph{arXiv preprint arXiv:2001.06057}, 2020.

\bibitem[Foret et~al.(2021)Foret, Kleiner, Mobahi, and
  Neyshabur]{foret2021sharpnessaware}
Pierre Foret, Ariel Kleiner, Hossein Mobahi, and Behnam Neyshabur.
\newblock Sharpness-aware minimization for efficiently improving
  generalization.
\newblock In \emph{International Conference on Learning Representations}, 2021.
\newblock URL \url{https://openreview.net/forum?id=6Tm1mposlrM}.

\bibitem[Keskar et~al.(2017)Keskar, Mudigere, Nocedal, Smelyanskiy, and
  Tang]{keskar2017on}
Nitish~Shirish Keskar, Dheevatsa Mudigere, Jorge Nocedal, Mikhail Smelyanskiy,
  and Ping Tak~Peter Tang.
\newblock On large-batch training for deep learning: Generalization gap and
  sharp minima.
\newblock In \emph{International Conference on Learning Representations}, 2017.
\newblock URL \url{https://openreview.net/forum?id=H1oyRlYgg}.

\bibitem[Jiang et~al.(2020)Jiang, Neyshabur, Mobahi, Krishnan, and
  Bengio]{Jiang2020Fantastic}
Yiding Jiang, Behnam Neyshabur, Hossein Mobahi, Dilip Krishnan, and Samy
  Bengio.
\newblock Fantastic generalization measures and where to find them.
\newblock In \emph{International Conference on Learning Representations}, 2020.
\newblock URL \url{https://openreview.net/forum?id=SJgIPJBFvH}.

\bibitem[Krizhevsky(2009)]{krizhevsky2009cifar}
Alex Krizhevsky.
\newblock Learning multiple layers of features from tiny images.
\newblock Technical report, 2009.

\bibitem[Le and Yang(2015)]{Le2015TinyIV}
Ya~Le and Xuan~S. Yang.
\newblock Tiny {ImageNet} visual recognition challenge.
\newblock 2015.

\bibitem[Deng et~al.(2009)Deng, Dong, Socher, Li, Li, and Fei-Fei]{deng2009}
Jia Deng, Wei Dong, Richard Socher, Li-Jia Li, Kai Li, and Li~Fei-Fei.
\newblock Imagenet: A large-scale hierarchical image database.
\newblock In \emph{IEEE Conference on Computer Vision and Pattern Recognition},
  pages 248--255, 2009.
\newblock \doi{10.1109/CVPR.2009.5206848}.

\bibitem[Zhang et~al.(2024)Zhang, Pan, Kim, Kweon, and Mao]{imagenet-d}
Chenshuang Zhang, Fei Pan, Junmo Kim, In~So Kweon, and Chengzhi Mao.
\newblock Imagenet-d: Benchmarking neural network robustness on diffusion
  synthetic object.
\newblock In \emph{Proceedings of the IEEE/CVF Conference on Computer Vision
  and Pattern Recognition (CVPR)}, pages 21752--21762, June 2024.

\bibitem[Hendrycks et~al.(2021)Hendrycks, Zhao, Basart, Steinhardt, and
  Song]{imagenet-a}
Dan Hendrycks, Kevin Zhao, Steven Basart, Jacob Steinhardt, and Dawn Song.
\newblock Natural adversarial examples.
\newblock In \emph{Proceedings of the IEEE/CVF Conference on Computer Vision
  and Pattern Recognition (CVPR)}, pages 15262--15271, June 2021.

\bibitem[Wang et~al.(2019)Wang, Ge, Lipton, and Xing]{imagenet-sketch}
Haohan Wang, Songwei Ge, Zachary Lipton, and Eric~P Xing.
\newblock Learning robust global representations by penalizing local predictive
  power.
\newblock In H.~Wallach, H.~Larochelle, A.~Beygelzimer, F.~d\textquotesingle
  Alch\'{e}-Buc, E.~Fox, and R.~Garnett, editors, \emph{Advances in Neural
  Information Processing Systems}, volume~32. Curran Associates, Inc., 2019.
\newblock URL
  \url{https://proceedings.neurips.cc/paper_files/paper/2019/file/3eefceb8087e964f89c2d59e8a249915-Paper.pdf}.

\bibitem[Salvador and Oberman(2022)]{imagenet-cartoon-drawing}
Tiago Salvador and Adam~M Oberman.
\newblock Imagenet-cartoon and imagenet-drawing: two domain shift datasets for
  imagenet.
\newblock In \emph{ICML 2022 Shift Happens Workshop}, 2022.
\newblock URL \url{https://openreview.net/forum?id=YlAUXhjwaQt}.

\bibitem[Taesiri et~al.(2023)Taesiri, Nguyen, Habchi, Bezemer, and
  Nguyen]{taesiri2023zoom}
Mohammad~Reza Taesiri, Giang Nguyen, Sarra Habchi, Cor-Paul Bezemer, and Anh
  Nguyen.
\newblock Imagenet-hard: The hardest images remaining from a study of the power
  of zoom and spatial biases in image classification.
\newblock 2023.

\bibitem[Goodfellow et~al.(2014)Goodfellow, Shlens, and Szegedy]{fgsm}
Ian~J Goodfellow, Jonathon Shlens, and Christian Szegedy.
\newblock Explaining and harnessing adversarial examples.
\newblock \emph{arXiv preprint arXiv:1412.6572}, 2014.

\bibitem[He et~al.(2016{\natexlab{b}})He, Zhang, Ren, and Sun]{he2016identity}
Kaiming He, Xiangyu Zhang, Shaoqing Ren, and Jian Sun.
\newblock Identity mappings in deep residual networks.
\newblock In \emph{European Conference on Computer Vision}, 2016{\natexlab{b}}.

\bibitem[Michaelis et~al.(2019)Michaelis, Mitzkus, Geirhos, Rusak, Bringmann,
  Ecker, Bethge, and Brendel]{michaelis2019dragon}
Claudio Michaelis, Benjamin Mitzkus, Robert Geirhos, Evgenia Rusak, Oliver
  Bringmann, Alexander~S. Ecker, Matthias Bethge, and Wieland Brendel.
\newblock Benchmarking robustness in object detection: Autonomous driving when
  winter is coming.
\newblock \emph{arXiv preprint arXiv:1907.07484}, 2019.

\bibitem[Srivastava et~al.(2014)Srivastava, Hinton, Krizhevsky, Sutskever, and
  Salakhutdinov]{srivastava14a}
Nitish Srivastava, Geoffrey Hinton, Alex Krizhevsky, Ilya Sutskever, and Ruslan
  Salakhutdinov.
\newblock Dropout: A simple way to prevent neural networks from overfitting.
\newblock \emph{Journal of Machine Learning Research}, 15\penalty0
  (56):\penalty0 1929--1958, 2014.
\newblock URL \url{http://jmlr.org/papers/v15/srivastava14a.html}.

\bibitem[Yuan et~al.(2021)Yuan, Chen, Wang, Yu, Shi, Jiang, Tay, Feng, and
  Yan]{yuan2021tokens}
Li~Yuan, Yunpeng Chen, Tao Wang, Weihao Yu, Yujun Shi, Zi-Hang Jiang,
  Francis~EH Tay, Jiashi Feng, and Shuicheng Yan.
\newblock Tokens-to-token vit: Training vision transformers from scratch on
  imagenet.
\newblock In \emph{Proceedings of the IEEE/CVF international conference on
  computer vision}, pages 558--567, 2021.

\bibitem[Wan et~al.(2013)Wan, Zeiler, Zhang, Le~Cun, and
  Fergus]{pmlr-v28-wan13}
Li~Wan, Matthew Zeiler, Sixin Zhang, Yann Le~Cun, and Rob Fergus.
\newblock Regularization of neural networks using dropconnect.
\newblock In Sanjoy Dasgupta and David McAllester, editors, \emph{Proceedings
  of the 30th International Conference on Machine Learning}, volume~28 of
  \emph{Proceedings of Machine Learning Research}, pages 1058--1066, Atlanta,
  Georgia, USA, 17--19 Jun 2013. PMLR.
\newblock URL \url{https://proceedings.mlr.press/v28/wan13.html}.

\bibitem[Kingma et~al.(2015)Kingma, Salimans, and
  Welling]{kingma2015variational}
Durk~P Kingma, Tim Salimans, and Max Welling.
\newblock Variational dropout and the local reparameterization trick.
\newblock \emph{Advances in neural information processing systems}, 28, 2015.

\bibitem[Graves(2011)]{graves2011}
Alex Graves.
\newblock Practical variational inference for neural networks.
\newblock In J.~Shawe-Taylor, R.~Zemel, P.~Bartlett, F.~Pereira, and K.Q.
  Weinberger, editors, \emph{Advances in Neural Information Processing
  Systems}, volume~24. Curran Associates, Inc., 2011.
\newblock URL
  \url{https://proceedings.neurips.cc/paper_files/paper/2011/file/7eb3c8be3d411e8ebfab08eba5f49632-Paper.pdf}.

\bibitem[Blundell et~al.(2015)Blundell, Cornebise, Kavukcuoglu, and
  Wierstra]{blundell2015weight}
Charles Blundell, Julien Cornebise, Koray Kavukcuoglu, and Daan Wierstra.
\newblock Weight uncertainty in neural network.
\newblock In \emph{International conference on machine learning}, pages
  1613--1622. PMLR, 2015.

\bibitem[Gal and Ghahramani(2016)]{pmlr-v48-gal16}
Yarin Gal and Zoubin Ghahramani.
\newblock Dropout as a bayesian approximation: Representing model uncertainty
  in deep learning.
\newblock In Maria~Florina Balcan and Kilian~Q. Weinberger, editors,
  \emph{Proceedings of The 33rd International Conference on Machine Learning},
  volume~48 of \emph{Proceedings of Machine Learning Research}, pages
  1050--1059, New York, New York, USA, 20--22 Jun 2016. PMLR.
\newblock URL \url{https://proceedings.mlr.press/v48/gal16.html}.

\bibitem[Louizos and Welling(2017)]{louizos2017multiplicative}
Christos Louizos and Max Welling.
\newblock Multiplicative normalizing flows for variational bayesian neural
  networks.
\newblock In \emph{International Conference on Machine Learning}, pages
  2218--2227. PMLR, 2017.

\bibitem[Izmailov et~al.(2021)Izmailov, Nicholson, Lotfi, and
  Wilson]{izmailov2021dangers}
Pavel Izmailov, Patrick Nicholson, Sanae Lotfi, and Andrew~G Wilson.
\newblock Dangers of bayesian model averaging under covariate shift.
\newblock \emph{Advances in Neural Information Processing Systems},
  34:\penalty0 3309--3322, 2021.

\bibitem[Ioffe and Szegedy(2015)]{batchnorm2015}
Sergey Ioffe and Christian Szegedy.
\newblock Batch normalization: accelerating deep network training by reducing
  internal covariate shift.
\newblock In \emph{Proceedings of the 32nd International Conference on
  International Conference on Machine Learning - Volume 37}, ICML'15, page
  448–456. JMLR.org, 2015.

\bibitem[Li et~al.(2016)Li, Wang, Shi, Liu, and Hou]{li2016revisiting}
Yanghao Li, Naiyan Wang, Jianping Shi, Jiaying Liu, and Xiaodi Hou.
\newblock Revisiting batch normalization for practical domain adaptation.
\newblock \emph{arXiv preprint arXiv:1603.04779}, 2016.

\bibitem[Nado et~al.(2020)Nado, Padhy, Sculley, D'Amour, Lakshminarayanan, and
  Snoek]{nado2020evaluating}
Zachary Nado, Shreyas Padhy, D~Sculley, Alexander D'Amour, Balaji
  Lakshminarayanan, and Jasper Snoek.
\newblock Evaluating prediction-time batch normalization for robustness under
  covariate shift.
\newblock \emph{arXiv preprint arXiv:2006.10963}, 2020.

\bibitem[Schneider et~al.(2020)Schneider, Rusak, Eck, Bringmann, Brendel, and
  Bethge]{schneider2020improving}
Steffen Schneider, Evgenia Rusak, Luisa Eck, Oliver Bringmann, Wieland Brendel,
  and Matthias Bethge.
\newblock Improving robustness against common corruptions by covariate shift
  adaptation.
\newblock \emph{Advances in neural information processing systems},
  33:\penalty0 11539--11551, 2020.

\bibitem[Benz et~al.(2021)Benz, Zhang, Karjauv, and Kweon]{benz2021revisiting}
Philipp Benz, Chaoning Zhang, Adil Karjauv, and In~So Kweon.
\newblock Revisiting batch normalization for improving corruption robustness.
\newblock In \emph{Proceedings of the IEEE/CVF winter conference on
  applications of computer vision}, pages 494--503, 2021.

\bibitem[B{\"o}hning and Lindsay(1988)]{bohning1988monotonicity}
Dankmar B{\"o}hning and Bruce~G Lindsay.
\newblock Monotonicity of quadratic-approximation algorithms.
\newblock \emph{Annals of the Institute of Statistical Mathematics},
  40\penalty0 (4):\penalty0 641--663, 1988.

\end{thebibliography}
